\documentclass{article} 
\usepackage{iclr2025_conference,times}

\usepackage{xcolor}
\definecolor{darkblue}{rgb}{0, 0.12, 0.55}
\definecolor{darkgreen}{rgb}{0, 0.55, 0.12}
\definecolor{darkred}{rgb}{0.6,0,0}
\definecolor{darkgreen}{rgb}{0,0.6,0}
\definecolor{Gray}{gray}{0.9}
\usepackage[breaklinks=true,
            colorlinks,
            linkcolor = darkred,
            urlcolor  = darkblue, 
            citecolor = teal,
            bookmarks = false]{hyperref}
\usepackage{url}
\usepackage{graphicx}
\usepackage{wrapfig}
\usepackage{booktabs}
\usepackage{adjustbox}
\usepackage{booktabs}
\usepackage{multirow}
\usepackage{xspace}
\usepackage{array}
\usepackage{bm}
\usepackage{bbm}
\usepackage[ruled,vlined]{algorithm2e}
\usepackage{amsthm,amsmath,amssymb}

\usepackage{listings}
\usepackage{wrapfig}
\usepackage{subcaption}
\usepackage{url}
\usepackage{colortbl}
\usepackage{adjustbox}
\usepackage{pifont}%
\definecolor{mymauve}{rgb}{0.58,0,0.82}

\author{Run Luo\textsuperscript{1,2}\thanks{Equal contribution. $\dag$ Min Yang and Binyuan Hui are corresponding authors.} 
\hspace{-1.1em}
\authorskip Yunshui Li\textsuperscript{1.2}$^*$ \hspace{-1.5em}
\authorskip Longze Chen\textsuperscript{1,2}$^*$ \hspace{-1.5em}
\authorskip Wanwei He\textsuperscript{1,2}
\hspace{-1.5em}
\authorskip Ting-En Lin\textsuperscript{3} \\ 
\hspace{-1.5em}
\textbf{\authorskip Ziqiang Liu\textsuperscript{1,2}} 
\textbf{Lei Zhang\textsuperscript{1,2}} \hspace{-1.5em}
\textbf{\authorskip Zikai Song\textsuperscript{4}} \hspace{-1.5em}
\textbf{\authorskip Hamid Alinejad-Rokny\textsuperscript{5}} 
\hspace{-1.5em}
\\
\hspace{-1.5em}
\textbf{\authorskip Xiaobo Xia\textsuperscript{6,7}} \hspace{-1.5em}
\textbf{\authorskip Tongliang Liu\textsuperscript{8}} 
\textbf{Binyuan Hui\textsuperscript{9}$^\dag$}
\hspace{-1.5em}
\textbf{\authorskip Min Yang\textsuperscript{1}$^\dag$} 
 \\[2mm]
\textsuperscript{1}Shenzhen Key Laboratory for High Performance Data Mining, SIAT, CAS \hspace{5.5mm} \\
\textsuperscript{2}University of Chinese Academy of Sciences \hspace{5.5mm} 
\textsuperscript{3}Tsinghua University \\
\textsuperscript{4}Huazhong University of Science and Technology 
\textsuperscript{5} University of New South Wales \\
\textsuperscript{6} School of Computing, National University of Singapore \\
\textsuperscript{7} MoE Key Laboratory of Brain-inspired Intelligent Perception and Cognition, \\ University of Science and Technology of China 
\textsuperscript{8} The University of Sydney 
\textsuperscript{9}Alibaba Group \hspace{5.5mm} 
}

\newcommand{\authorskip}{\hspace{4.8mm}}

\newcommand{\resolved}[1]{}

\newcommand{\com}[1]{}

\newlength{\mysize}

\newcommand{\cmark}{{\ding{51}}}%
\newcommand{\xmark}{{\ding{55}}}%
\definecolor{mydarkblue}{rgb}{0,0.08,0.55}
\definecolor{drp-blue}{HTML}{1f77b4}
\definecolor{mygray}{gray}{.9}
\definecolor{light-gray}{gray}{0.5}
\definecolor{pretty-blue}{RGB}{0, 113, 188}
\definecolor{maroon}{RGB}{0,128,128}
\definecolor{linecolor1}{gray}{.95} 
\definecolor{linecolor}{gray}{.895} 
\definecolor{kaiming-green}{RGB}{57,181,74} 
\definecolor{icmlblue}{rgb}{0,0.08,0.45} 
\definecolor{prompt_blue}{HTML}{1f78b4}
\definecolor{prompt_red}{HTML}{d45c43}

\def\ie{{\it{i.e.}}}

\usepackage[capitalize]{cleveref} 

\def\name{DEEM }

\iclrfinalcopy 

\title{\name: \underline{D}iffusion Models Serve as the \underline{E}y\underline{E}s of Large Language \underline{M}odels for Image Perception}

\begin{document}

\maketitle

\begin{abstract}\label{sec:abs}
The development of large language models (LLMs) has significantly advanced the emergence of large multimodal models (LMMs). While LMMs have achieved tremendous success by promoting the synergy between multimodal comprehension and creation, they often face challenges when confronted with out-of-distribution data, such as which can hardly distinguish orientation, quantity, color, structure, etc. This is primarily due to their reliance on image encoders trained to encode images into task-relevant features, which may lead them to disregard irrelevant details. Delving into the modeling capabilities of diffusion models for images naturally prompts the question: Can diffusion models serve as the eyes of large language models for image perception? In this paper, we propose \name, a simple but effective approach that utilizes the generative feedback of diffusion models to align the semantic distributions of the image encoder. This addresses the drawbacks of previous methods that solely relied on image encoders like CLIP-ViT, thereby enhancing the model's resilience against out-of-distribution samples and reducing visual hallucinations. Importantly, this is achieved without requiring additional training modules and with fewer training parameters. We extensively evaluated \name on both our newly constructed RobustVQA benchmark and other well-known benchmarks, POPE and MMVP, for visual hallucination and perception. In particular, \name improves LMM's visual perception performance to a large extent (e.g., 4\% ↑ on RobustVQA, 6.5\% ↑ on MMVP, and 12.8 \% ↑ on POPE ). Compared to the state-of-the-art interleaved content generation models, \name exhibits enhanced robustness and a superior capacity to alleviate model hallucinations while utilizing fewer trainable parameters, less pre-training data (10\%), and a smaller base model size. Extensive experiments demonstrate that \name enhances the performance of LMMs on various downstream tasks without inferior performance in the long term, including visual question answering, image captioning, and text-conditioned image synthesis. The code and benchmark are available at 
\url{https://github.com/RainBowLuoCS/DEEM}
\end{abstract}

\begin{figure*}
\centering
\includegraphics[width=0.96\textwidth]{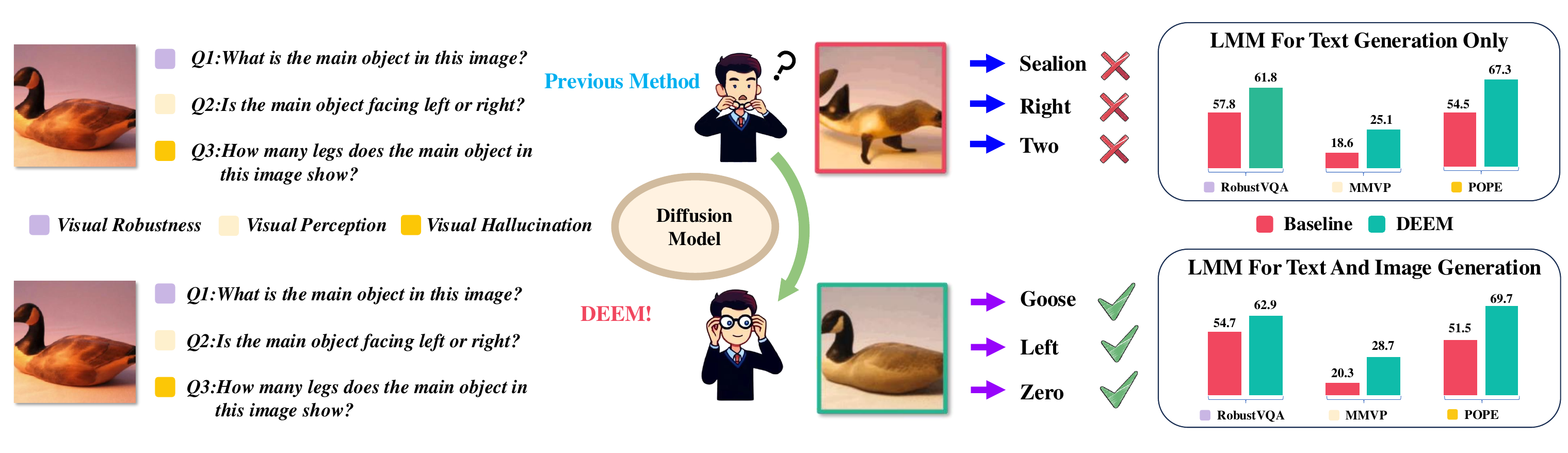}
\captionof{figure}{\textbf{Illustration of our \name.} When encountering natural adversarial examples or out-of-distribution data, DEEM uses the diffusion model to check if the semantic features of the image encoder match the input images. This approach allows DEEM to serve as the "eyes" of the large language model, proactively identifying and correcting misinterpreted semantic information during training, thereby avoiding the loss of important visual details. This enhances the robustness, hallucination recognition, and foundational visual perception capabilities of LMMs. In contrast, other models rely too heavily on erroneous inputs from the image encoder, making it difficult for them to handle challenges posed by such data.}
\label{fig:overview}
\end{figure*}

\section{Introduction}\label{sec:intro}
\vspace{-4pt}
With the success of large language models (LLMs), large multimodal models (LMMs) built on LLMs have garnered significant attention. Researchers~\citep{llava,minigpt4,instructblip, flamingo, shikra} have attempted to build a bridge between large language models and image encoders through simple mapping modules, and have already made significant progress in multimodal understanding tasks such as visual question answering. Subsequent studies~\citep{cm3leon,emu,dreamllm,mm-interleaved} utilize extra advanced diffusion models (DMs)~\citep{stablediffusion} for image generation and train the LMMs on interleaved text-image data in an end-to-end manner. This unified paradigm of multimodal understanding and creation brings various isolated multimodal tasks together, greatly boosting model capabilities and expanding application scenarios.

However, these models commonly rely on encoder architectures like CLIP-ViT~\citep{clip}, which suffers from certain perceptual understanding
limitations due to the contrastive learning paradigm and the noisy image-text pairs used in training, to encode input images. Additionally, these image encoders are typically trained to encode images into features relevant to downstream tasks, thereby disregarding irrelevant details. Consequently, as shown in \cref{fig:overview}, when faced with images outside the training scope, they often capture biased semantic features, resulting in erroneous visual information being perceived by subsequent language models. This accumulation of inaccuracies renders the multimodal model unable to comprehend multimodal context effectively. For this reason, this makes it difficult for previous methods to discern subtle details, thereby hindering their ability to handle tasks related to basic visual perception, visual hallucinations, and visual robustness that are very simple for humans.

On the contrary, the goal of diffusion models~\citep{diffusionmodel} is to learn a diffusion process that characterizes a probability distribution for a given dataset, without direct training on the downstream task objective. This enables it to capture finer details of images for better handling of out-of-distribution data. However, there have been few efforts to integrate the capabilities of the diffusion model into the image perception of large multimodal models.

In this paper, we propose \textbf{DEEM}, a simple but effective approach to leverage the generative feedback of diffusion models for aligning the semantic distributions of image encoders in an elegant self-supervised manner. Building upon this, we introduce an end-to-end interleaved image-text generative modeling approach, where diffusion models serve as additional eyes of large language models for image perception. This addresses the limitations of previous methods that solely relied on image encoders such as CLIP-ViT~\citep{clip}, enhancing the model's robustness against out-of-distribution samples and reducing hallucination perception in multimodal scenarios, without the need for additional training modules and with fewer training parameters. To the best of our knowledge, we are the first to apply diffusion models to large multimodal models for image perception.

Specifically, DEEM takes interleaved image-text pairs as input to the model. It starts by encoding images and text using corresponding visual and text encoders, resulting in image tokens and text tokens. These tokens are then organized according to their original layout and inputted into a large language model to generate corresponding hidden state outputs. The model employs autoregressive modeling for the hidden state outputs of text and utilizes the output hidden states of images, along with the image tokens encoded by the image encoder, as diffusion conditions. These conditions are then fed into a diffusion model for image reconstruction. Through end-to-end training, the model not only acquires the capacity to generate text and images but also employs semantic consistency regularization on the semantic information produced by the image encoder during image reconstruction. This compels the image encoder to incorporate more details into the semantic representation of the image, thereby mitigating the issue of semantic bias in image encoding.

DEEM is trained on a mixture corpora of image-text pairs and interleaved image-text sequences data without extra in-house data following previous solution~\citep{blip,blip2,dreamllm,mm-interleaved}. To assess the robustness recognition capability of LMMs, we constructed a new robustness benchmark, RobustVQA, based on existing datasets containing natural adversarial samples and out-of-distribution data. RobustVQA is divided into three parts: RobustVQA-A, RobustVQA-R, and RobustVQA-V, based on different data sources, aiming to provide better insights into the performance of LMMs in real-world scenarios. We conducted extensive evaluations of DEEM on RobustVQA and two widely recognized benchmarks, POPE and MMVP, for visual hallucination and perception respectively.
Experimental results indicate that our method exhibits enhanced robustness, a superior capacity to alleviate model hallucinations and better visual perception ability in comparison to the state-of-the-art interleaved image-text modeling model MM-Interleaved~\citep{mm-interleaved}, using a smaller-scale image encoder (CLIP-ConvNext-B~\citep{convnext} vs. CLIP-ViT-L~\citep{clip}), a smaller-scale language model (Vicuna 7B vs. Vicuna 13B~\citep{vicuna}), and less pre-training data (without Laion-coco~\citep{laion-coco} \& Laion-en~\citep{laion-en}). \name outperforms MM-Interleaved 9.4\% on RobustVQA, 17.8\% on POPE and 9.1\% on MMVP. Moreover, with further enhancement via supervised fine-tuning, \name achieves competitive results on various multimodal tasks, including visual question-answering, region-level image captioning, and text-to-image generation.

Before delving into details, we summarize our contributions as follows.

 \ $\bullet$ \textbf{ Robustness Benchmark.} We design a new robustness benchmark RobustVQA for LMMs based on publicly available ImageNet-A~\citep{imagenet-a}, ImageNet-R~\citep{imagenet-r}, and ImageNet-V2~\citep{imagenet-v2} datasets, which can be utilized to effectively assess the visual robustness capabilities of the multimodal models.

 \ $\bullet$ \textbf{ Effective Method. } We are the first to introduce the diffusion model into the image perception of large language models, to correct potential semantic bias in the image encoder and alleviate the excessive compression of visual details. This approach enhances the model's robustness and hallucination mitigation capabilities without the need for additional modules or trainable parameters.

\ $\bullet$ \textbf{ DEEM Model.} Based on the proposed method, we train a multimodal model with end-to-end interleaved text-image modeling capabilities. After supervised fine-tuning, \name can perform various multimodal tasks in a unified manner, such as visual question answering, text-to-image generation, and region-level image captioning.

\ $\bullet$ \textbf{ Comprehensive Experiments.} We provide abundant qualitative and quantitative comprehensive experimental results to demonstrate the effectiveness and efficiency of the proposed method.

\section{Method}\label{sec:method}
\vspace{-4pt}
In this section, we first present our \name, starting with an introduction to the overall architecture in \cref{subsec:arch}, followed by a description of the pipeline in \cref{subsec:pipe}. Finally, we provide details on the training and inference process in \cref{subsec:training}.

\subsection{Architecture}\label{subsec:arch}
In this subsection, we present the multi-modal architecture for processing interleaved
image-text data. To excel in both comprehension and creation tasks of text and images, a multi-modal model consists of the following three key components. 

\textbf{VFM-based Image Encoder $\mathcal{E}_V$} which encodes each image $x^V \in \mathbb{R}^{H \times W \times 3}$ into an image embedding $e^V \in \mathbb{R}^{N \times C}$, where $C$ is the channel dimension and $N$ is the number of visual tokens in image embedding. \textbf{LLM-based Multi-modal Decoder $\mathcal{D}_\text{LLM}$} that extracts context features from the interleaved image-text token sequences. Its input sequence $E\in \mathbb{R}^{K \times C}$ is a concatenation of embeddings $(e_1, e_2, \dots)$, where $e_n$ is either a word embedding $e_n^L \in \mathbb{R}^{1 \times C}$ or an image embedding $e_n^V \in \mathbb{R}^{N \times C}$. $K$ is the total number of input tokens. \textbf{DM-based Image Decoder $\mathcal{D}_\text{DM}$} that generates the image conditioned on image-text sequences context feature. 

To provide the conditional inputs for $\mathcal{D}_\text{DM}$ and reduce the number of visual tokens in image embedding $e^V$, two different Perceiver Resampler~\citep{flamingo} are employed to map the output features from multi-modal decoder $\mathcal{D}_{\text{LLM}}$ and image encoder $\mathcal{E}_{V}$ to a fixed number of conditional tokens, respectively. Additionally, we utilize an extra mask-aware visual extractor $\mathcal{E}_\text{M}$ for extracting region visual information from image embedding $e^V$ via simple mask-aware operation $\mathcal{E}_\text{M}(e^V, \mathcal{M}^V)$, where $\mathcal{M}^V$ is the corresponding binary mask of image $x^V$. 

\begin{figure*}[t!]
    \centering
    \includegraphics[width=\linewidth]{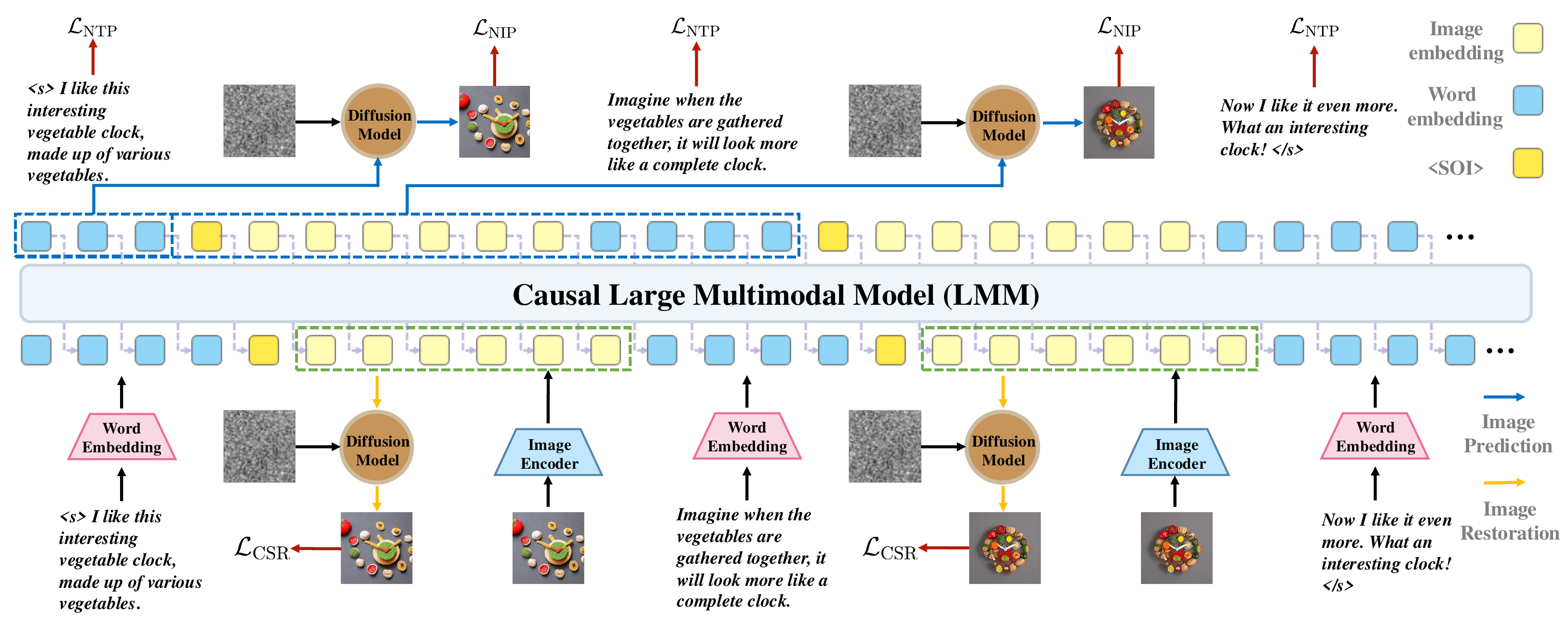}
    \vspace{-0.55cm}
    \caption{\textbf{Overview of our \name framework}. Interleaved documents serve as input, decoded to produce outputs. Both text and images are encoded into sequential, discrete token embeddings for the LMM input. Here, we replace the $<$IMG$>$ token embedding in the text with the image embedding before inputting it into the LLM. The text is predicted in an autoregressive manner and the images are synthesized by the DM-based image decoder conditioned on holistic historical semantics captured by LMM. Besides, the image token embeddings are fed into DM-based image decoder for consistent image restoration. The start of image token $<$SOI$>$ is used to determine the starting position of the image, facilitating the natural autoregressive generation of interleaved text-image layouts. Note that our core architecture is presented without the connectors between modules for simplicity.}
\label{fig:framework}
\vspace{-0.65cm}
\end{figure*}

\subsection{Pipeline}\label{subsec:pipe}
\begin{wrapfigure}{r}{0.45\textwidth}
    \vspace{-0.4cm}
    \centering
    \includegraphics[width=\linewidth]{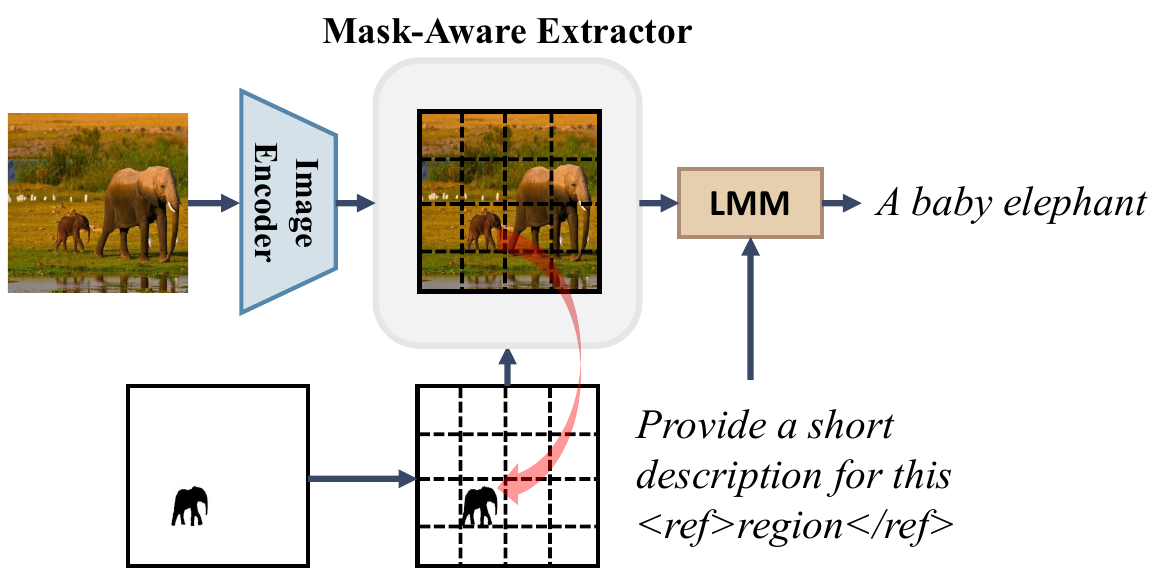}
    \vspace{-0.58cm}
    \caption{
    \textbf{Pipeline of Mask-Aware Extractor}. The mask-aware extractor can be used to extract region-level visual features based on the mask-aware operation. A simple dot product is applied between the mask and the image embedding before being fed into the LLM.
 }\label{fig:mask_aware_extractor}
\vspace{-0.50cm}
\end{wrapfigure}

As shown in \cref{fig:framework}, given an interleaved image-text sequence $X=\{x_1, x_2, x_3, \dots\}$, where each element $x_n$ is either a text token (denoted as $x_n^L$) or a whole image (denoted as $x_n^V$). Text and images are arranged in the order in which they appear in the original content. 
To build an end-to-end generative model for interleaved image-text data, a common practice is to first extract embedding for each text token and each image and then feed them into LLMs, \ie, $e_n^L = \mathcal{E}_L(x_n^L)$ and  $e_n^V = \mathcal{E}_\text{M}(\mathcal{E}_V(x_n^V), \mathcal{M}_n^V)$, where $\mathcal{E}_L$ denotes word embedding in LLM. $\mathcal{E}_V$ is typically an image encoder followed by a Perceiver Resampler~\citep{flamingo} to map each image to a fixed number of visual tokens. As shown in \cref{fig:mask_aware_extractor}, we introduce a mask-aware visual extractor $\mathcal{E}_\text{M}$ for extracting region visual information from image embedding $e_n^V$ via simple mask-aware operation $\mathcal{E}_\text{M}(e_n^V, \mathcal{M}_n^V)$, where $\mathcal{M}_n^V$ is the corresponding binary mask of image $x_n^V$ and the default value is 1. Then, the interleaved generative modeling is trained to maximize the log-likelihood:
\begin{equation}
    \begin{aligned}
        \log p(X) = \sum_n \log p(x_n | e_{< n})
        = \sum_{n \in \mathcal{I}_L}\underbrace{\log p(x_n^L | e_{< n})}_{\text{text ~ prediction}} + \sum_{n \in \mathcal{I}_V} \underbrace{\log p(x_n^V | e_{< n})}_{\text{image ~ prediction}},
    \end{aligned}
    \label{equ:formu}
\end{equation}
where $\mathcal{I}_L$ and $\mathcal{I}_V$ represent the index sets for text tokens and images, respectively. That $<n$ in the subscript represents the abbreviation of $\{1, 2, \dots, n-1\}$. The following paragraphs provide explanations of Eq.~\eqref{equ:formu}.

\textbf{Text Generation with Multi-modal Condition.}
$\log p(x_n^L | e_{< n})$ is similar to traditional causal language modeling, except that the condition also includes previous images. 
Recent works~\citep{flamingo,blip2,llava} have demonstrated the effectiveness of using LLMs for processing additional visual inputs. 
The loss function for text generation is
\begin{equation}
    \begin{aligned}
        \mathcal{L}_\text{NTP}(x_n^L | e_{< n}) = - \log p(x_n^L | \mathcal{D}_\text{LLM}(e_{< n}) \big),
    \end{aligned}
    \label{equ:loss_text}
\end{equation}
where $\mathcal{D}_\text{LLM}$ denotes the LLM network.

\textbf{Image Generation with Multi-modal Condition.}
Maximizing $\log p(x_n^V | e_{< n})$ aligns with the diffusion denoising process, which recently achieved widespread success in image generation. Maximizing the log-likelihood is derived as minimizing the diffusion modeling loss as
\begin{equation}
    \begin{aligned}
        \mathcal{L}_\text{NIP}(x_n^V | e_{< n}) = \mathbb{E}_{\epsilon,t}~||\epsilon - \mathcal{D}_\text{DM}\big(x_{n,t}^V, t, \mathcal{D}_\text{LLM}(e_{< n})\big)||^2,
    \end{aligned}
    \label{equ:loss_img}
\end{equation}
where $\mathcal{D}_{DM}$ is the diffusion model for denoising process. That $x_{n,t}^V$ is the noisy version of the original image at the denoising step $t$, and the denoising network $\mathcal{D}_{DM}$ is trained to predict the noise $\epsilon$.

\textbf{Consistency Semantic Regularization.} In addition to the above text and image generation loss functions, we propose a new consistency semantic constraint term. This term reuses the diffusion model to perform generative checks on the image semantic information extracted by the image encoder, ultimately correcting erroneous knowledge in the pre-trained image encoder. This significantly enhances the out-of-distribution generalization and reduces visual hallucinations in the multi-modal model. The new log-likelihood function can be written as
\begin{equation}
    \begin{aligned}
        \log p^\star(X) = \sum_{n \in \mathcal{I}_L} \underbrace{\log p(x_n^L | e_{< n})}_{\text{text~prediction}} + \sum_{n \in \mathcal{I}_V} \underbrace{\log p(x_n^V | e_{< n})}_{\text{image~prediction}}+
        \sum_{n \in \mathcal{I}_V} \underbrace{\log p(x_n^V | e_n)}_{\text{image~restoration}}.
    \end{aligned}
    \label{equ:new_formu}
\end{equation}
Similarly, the corresponding log-likelihood function $\log p(x_n^V | e_n)$ can be equivalently written as the following loss function used in training:
\begin{equation}
    \begin{aligned}
        \mathcal{L}_\text{CSR}(x_n^V | e_n) = 
        \mathbb{E}_{\epsilon,t}~||\epsilon - \mathcal{D}_\text{DM}\big(x_{n,t}^V, t, e_n\big)||^2.
    \end{aligned}
    \label{equ:loss_img_regular}
\end{equation}

Note that the new end-to-end modeling framework brings significant improvements to the generalization performance of the model without altering the original modeling flexibility or introducing additional modules.

\subsection{Training and Inference}\label{subsec:training}
We employ a three-stage training process, consisting of image-text alignment pre-training, image-text instruction fine-tuning, and mask-text instruction fine-tuning. Image-text alignment pre-training and image-text instruction fine-tuning are designed to validate the effectiveness and efficiency of semantic consistency regularization in enhancing the visual perception capabilities of LMMs. Mask-text instruction fine-tuning is used to verify whether the model trained with semantic consistency regularization negatively impacts the performance of fine-tuning on downstream tasks in the long term. The image-text alignment pre-training objective is defined as the sum of the next-text prediction loss in Eq.~\eqref{equ:loss_text}, next-image prediction loss in Eq.~\eqref{equ:loss_img} and consistency semantic regularization loss in Eq.~\eqref{equ:loss_img_regular} as $\mathcal{L}_{S_1} =  \mathcal{L}_{\text{NTP}} + \lambda~\mathcal{L}_{\text{NIP}}+\lambda~\mathcal{L}_{\text{CSR}}$,
where $\lambda$ is a coefficient used to determine the relative loss weight between the image and text decoding branches. In order to enable the \name to perform general multimodal comprehension and creative tasks following human instructions, we use $\mathcal{L}_{S_2} =  \mathcal{L}_{\text{NTP}} +\lambda~\mathcal{L}_{\text{CSR}}$ to conduct image-text instruction fine-tuning. To further enhance the model's fine-grained region awareness, we conducted region-level mask-text instruction fine-tuning. Since there is no need to perform text-to-image tasks, we removed the next-image prediction loss and the training objective in mask-text instruction fine-tuning can be defined as $\mathcal{L}_{S_3} =  \mathcal{L}_{\text{NTP}}$. 
 The whole framework can be optimized end-to-end during the three stages. During inference, the images and texts are generated in an auto-regressive manner. Text tokens are sampled from the distribution predicted by the multi-modal LLM. When the generated token is \verb|<SoI>|, the diffusion model is called for generating the next image.

\section{Experiment}\label{sec:exp}
\vspace{-4pt}
\subsection{Implementation Details}
In this subsection, we first introduce the network of \name and then showcase the three-stage training recipes. More details of datasets and hyper-parameters can be found in \cref{tab:details}.

\textbf{Network.} Similar to previous work, We leverage Vicuna7B~\citep{vicuna} and Stable Diffusion v2.1~\citep{stablediffusion} as the large language model, and image decoder, respectively. However, unlike their use of a 427M parameter CLIP-ViT-L as the image encoder, we use a smaller 122M parameter CLIP-ConvNeXt-B\citep{convnext}. For the multi-modal LLM, two different Perceiver Resamplers~\citep{flamingo} are used to connect diffusion model with image encoder and large language model respectively. 

\textbf{Image-Text Alignment Pre-training.} Our model is pre-trained on a mixture of image-text pairs and interleaved image-text sequences, including MMC4-Core~\citep{mmc4}, LAION-400M~\citep{laion-400m}, SBU~\citep{sbu}, and CC-12M~\citep{cc}. For LAION-400M~\citep{laion-400m}, SBU~\citep{sbu}, and CC-12M~\citep{cc}, instead of utilizing the original annotations, we use the version filtered by the pre-trained BLIP-2 model~\citep{blip2}. For simplicity, we refer to it as BLIP-LCS hereafter. "LCS" abbreviates the LAION, CC, and SBU datasets. The sampling probability of MMC4 is twice that of BLIP-LCS. The images are inserted before or after the corresponding text sentence with equal probability. To optimize training efficiency and data utility, multiple image-text pairs or interleaved image-text sequences are concatenated into extended sequences with the maximum context length.

\textbf{Image-Text Instruction Fine-tuning.} 
To enable \name to perform general multimodal comprehension tasks following human instructions, we utilize publicly available datasets for image-text instruction fine-tuning, including LLaVA-665K~\citep{llava}, COCO
Caption~\citep{coco}, VQAv2~\citep{vqav2}, TextCaps~\citep{textcaps}, OCR-VQA~\citep{ocrvqa}, GQA~\citep{gqa}, OK-VQA~\citep{ok-vqa}, TextVQA~\citep{textvqa}, and
AOK-VQA~\citep{aokvqa}.

\begin{wrapfigure}{r}{0.5\textwidth}
    \vspace{-0.48cm}
    \centering
    \includegraphics[width=\linewidth]{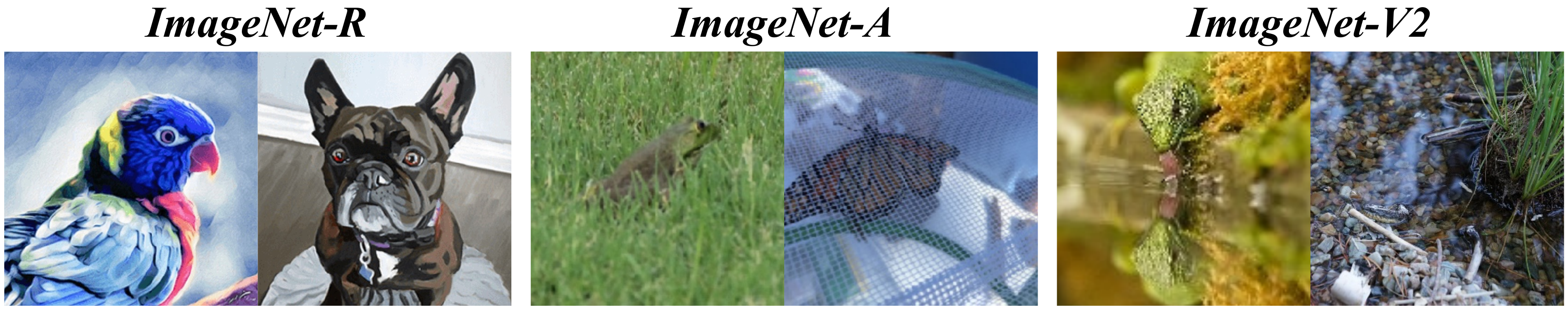}
    \vspace{-0.58cm}
    \caption{
    \textbf{Examples from ImageNet-R, ImageNet-A, and ImageNet-V2}. These examples share similar backgrounds, rare materials, and unusual textures. They serve as natural adversarial examples and out-of-distribution data, which can be used to test the robustness of models.
 }\label{fig:robust_example}
\vspace{-0.48cm}
\end{wrapfigure}

\textbf{Mask-Text Instruction Fine-tuning.}  At this stage, we use a simple mask-aware visual extractor to capture pixel-level region features and then align mask-based region features with language embeddings. We collect short text and pixel-level mask pairs from the publicly available object-level datasets (COCO~\citep{coco}, RefCOCO~\citep{refercoco},
RefCOCO+~\citep{refercoco+g}, RefCOCOg~\citep{refercoco+g}), part-level datasets (Pascal Part~\citep{pascalpart}, Part Imagenet~\citep{partimagenet}), and multiple region datasets(VCR~\citep{vcr}, Visual Genome~\citep{vg}). Then we conduct mask-text instruction fine-tuning on the mixture of the above text-mask pairs data, enabling \name to complete region-level understanding tasks, such as region-level image captioning.

\subsection{Experimental Results}
\vspace{-0.25cm}
In this study, we evaluate our \name model by comparing it with current state-of-the-art (SOTA) models on various tasks including visual robustness , hallucination diagnosis, basic visual perception and image-level visual question answering. Please refer to \cref{app:add_exp} for more experimental results about mask-level visual question answering and text-to-image generation. All metrics and data splits are listed in \cref{tab:details} in \cref{app:imple_details}.

\begin{table*}[!b]
\centering
\footnotesize
\setlength\tabcolsep{2pt}
\caption{ \textbf{Zero-shot visual robustness, hallucination and perception evaluation} of RobustVQA-A: RVQA-A, RobustVQA-R: RVQA-R, RobustVQA-V: RVQA-V, POPE-Random: POPE-R~\citep{pope}, POPE-Popular: POPE-P~\citep{pope}, POPE-Adversarial: POPE-A~\citep{pope} 
and MMVP~\citep{mmvp} benchmarks. RobustVQA-A, RobustVQA-R, and RobustVQA-V are robustness benchmarks designed by us in \cref{subsec:dataset_construction}. "AVG" denotes the overall average accuracy of seven benchmarks. "SFT" denotes the supervised fine-tuning. "*" denotes baseline model without diffusion feedback. The evaluation metrics for each benchmark are listed in \cref{tab:eval_benchmarks_summary}.}\label{tab:robustness}
\resizebox{1.0\linewidth}{!}{
\begin{tabular}{lll|cccccccc}
\toprule

Method & SFT & Architecture & RVQA-A & RVQA-R & RVQA-V & POPE-R & POPE-P & POPE-A & MMVP & AVG \\
    \midrule[0.6pt]
    \multicolumn{11}{c}{\textit{Models for Text-Generation Only}} \\ 
    \midrule[0.6pt]
    Shikra~\citep{shikra} &\cmark &ViT-L/LLaMA 7B & 33.71 & 38.33 & 37.45 & 86.90 & 83.97 & 83.10 & 22.56 & 55.15\\
    NeXT-Chat~\citep{nextchat} & \cmark& ViT-L/Vicuna 7B & 44.82 & 43.67 & 47.30 & 87.70 & 84.57 & 81.93 & 27.41 & 59.62\\
    \midrule[0.6pt]
    \multicolumn{11}{c}{\textit{Models for Text and Image Generation}} \\ 
    \midrule[0.6pt]
    MM-Interleaved~\citep{mm-interleaved}& \xmark &ViT-L/Vicuna 13B & 50.76 & 52.71 & 50.60 & 64.73 & 65.33 & 65.20 & 23.82 & 53.31\\
    Emu-I~\citep{emu} & \cmark &ViT-L/Vicuna 7B & 46.40 & 49.12 & 47.36 & 61.28 & 56.79 & 56.01 & 22.69 & 48.52\\
    SEED~\citep{seed} & \cmark &ViT-G/Vicuna 7B & 52.06 & 59.71 & 57.02 & 69.84 & 56.83 & 59.63 & 25.62 & 54.39 \\
    DreamLLM~\citep{dreamllm} & \cmark & ViT-L/Vicuna 7B & 51.43 & 58.96 & 57.60 & 86.36 & 80.07 & 72.63 & 26.37 & 61.84\\
    SEED-X~\citep{seed-x} & \cmark &ViT-G/Vicuna 13B & 52.36 & 60.27 & 59.49 & 86.41 & 81.43 & 74.56 & 29.16 & 63.39\\
    \midrule[0.45pt]
    \name* &\xmark &ConvNext-B/Vicuna 7B & 53.24 & 56.06 & 54.72 & 50.55 & 52.00 & 51.93 & 20.30 & 48.40\\
    \name &\xmark& ConvNext-B/Vicuna 7B & \textbf{56.86} & \textbf{68.63} & \textbf{63.08} & 69.93 & 70.27 & 68.87 & 28.74 & 60.91 \\
    \name-VQA &\cmark& ConvNext-B/Vicuna 7B & 55.22 & 64.12 & 62.99 & \textbf{87.40} & \textbf{82.80} & \textbf{78.49} & \textbf{32.89} & \textbf{65.56}\\
    \bottomrule
\end{tabular}}
\vspace{-15pt}
\end{table*}

\textbf{Visual Perception Diagnose.} We explore the impact of diffusion feedback on the visual perception capabilities of LMMs from three dimensions: visual robustness, visual hallucinations, and basic visual perception. To rigorously assess visual robustness of our model, we design a benchmark called RobustVQA for robustness evaluation based on online datasets, including ImageNet-A~\citep{imagenet-a}, ImageNet-R~\citep{imagenet-r} and ImageNet-V2 ~\citep{imagenet-v2}. As shown in \cref{fig:robust_example}, these challenging natural adversarial examples and out-of-distribution samples in the original ImageNet dataset can be used to evaluate the neural network robustness of our model. Similar to the POPE and MMVP dataset, we first choose the challenging sample from ImageNet-A, ImageNet-R, and ImageNet-V2 dataset and then convert the them into a VQA format that the multimodal model can evaluate simply and accurately. More details about the new benchmark RobustVQA design can be found in \cref{subsec:dataset_construction}. 
For a comprehensive visual robustness and hallucination evaluation, we evaluate our model against other open-source state-of-the-art (SOTA) LMMs for text and image generation, including SEED~\citep{seed}, SEED-X~\citep{seed-x}, MM-Interleaved~\citep{mm-interleaved}, and DreamLLM~\citep{dreamllm}, on the RobustVQA, POPE ~\citep{pope} and MMVP ~\citep{mmvp} dataset with accuracy metric. The results, presented in \cref{tab:robustness}, demonstrate that our \name model not only exhibits competitive performance compared with existing fine-tuned SOTA models on POPE and MMVP after fine-tuning, but also achieves the best results among visual robustness benchmark only after pre-training. Notably, compared to the larger-scale concurrent SOTA model for interleaved text-image modeling, MM-Interleaved~\citep{mm-interleaved}, our model achieves better results with a smaller scale. DEEM outperforms MM-Interleaved 9.4\% on RobustVQA,
17.8\% on POPE and 9.1\% on MMVP. To ensure a fair comparison and prove the effectiveness of our method, we also train an MM-Interleaved model with the same experimental setting as a baseline. Compared to this baseline, Our method achieves an 4\% average gain on RobustVQA,
12.8\% average gain on POPE and 6.5\% average gain on MMVP, respectively. The experimental results demonstrate the effectiveness of our method for better LMMs' visual perception capability.

\begin{table*}[!h]\small
\centering
\footnotesize
\setlength\tabcolsep{2pt}
\caption{\textbf{Multi-modal comprehension evaluation}. ``ED'' denotes using extra in-house data. Benchmarks include COCO~\citep{coco}; I2Para.: Image2Paragraph~\citep{img2paragraph}; VQA$^{\rm v2}$: VQAv2~\citep{vqav2}; OKVQA~\citep{ok-vqa}; GQA~\citep{gqa}; VizWiz~\citep{vizwiz}; VisDial~\citep{visdial}; MMBench: MMB ~\citep{mmvet}; MMVet~\citep{mmvet};. The evaluation metrics for each benchmark are listed in \cref{tab:eval_benchmarks_summary}. }
\resizebox{0.95\linewidth}{!}{
\begin{tabular}{@{}lllc|cc|ccccc|cc@{}}
\toprule

Model & LLM &VFM & ED & COCO & I2Para. & VQA$^{\rm v2}$ & OKVQA & GQA & VizWiz & VisDial & MMB & MMVet \\ 
\midrule[0.6pt]
\multicolumn{13}{c}{\textit{Models for Text-Generation Only}} \\ 
\midrule[0.6pt]
IDEFICS-80B~\citep{idefics} & LLaMA-65B &ViT-H&\xmark&
91.8  & -- & 60.0 & -- & 45.2 & 36.0 & -- & 27.9 & -- \\

IDEFICS-80B-I~\citep{idefics} & LLaMA-65B &ViT-H&\xmark&
117.2 & -- & 37.4 & -- & -- & 26.0 & -- & --  & -- \\

KOSMOS-1~\citep{kosmos1} & MetaLM &ViT-L& \cmark & 
-- & -- & 
46.7 & -- & -- & -- &  -- & -- & --  \\

KOSMOS-2~\citep{kosmos2} & KOSMOS-1 &ViT-L&\cmark & 
--  & -- &
45.6 & -- & -- & -- &  --  & -- & -- \\

Flamingo-9B~\citep{flamingo} & Chinchilla-7B &ViT-L&\cmark& 
79.4 &  -- &
51.8 & 44.7 & -- & 28.8 &  48.0 & 7.9 & 23.3  \\

Flamingo-80B~\citep{flamingo} & Chinchilla-70B &ViT-H&\cmark& 
84.3 & -- &
56.3 & 50.6 & -- & 31.6 &  52.0 & -- & --  \\

mPLUG-DocOwl~\citep{mplug} & LLaMA-7B &ViT-L&\xmark & 
52.6 & -- & -- & -- & -- & --  & -- & 60.8 & 35.7 \\

BLIP-2~\citep{blip2} & Vicuna-7B &ViT-L&\xmark &
--  & -- & -- & -- & 38.6 & 25.3  & -- & -- & --  \\

BLIP-2~\citep{blip2} & Vicuna-13B &ViT-L&\xmark &
-- & -- & 41.0 & -- & 41.0 & 19.6  & -- & -- & --  \\

InstructBLIP~\citep{instructblip} & Vicuna-7B &ViT-L&\xmark &
--  & -- & --& -- & 49.2 & 34.5  & -- & 68.9 & 33.1  \\

InstructBLIP~\citep{instructblip} & Vicuna-13B & ViT-L&\xmark &
--& -- & --& -- & 49.5& 33.4 & -- & -- & -- \\

Shikra~\citep{shikra} & Vicuna-13B &ViT-L&\xmark &
117.5 & -- &77.4 & -- & -- & --  & -- & -- & --  \\

LLaVA-1.5~\citep{llava2} & Vicuna-7B &ViT-L&\xmark 
 & --  & -- & 78.5 & -- & 62.0 & 50.0  & -- & 53.1 & 32.9   \\

LLaVA-1.5~\citep{llava2} & Vicuna-13B &ViT-L&\xmark & -- & -- & 80.0 & -- & 63.3 & 53.6  & -- & 60.6 & 35.6  \\

Qwen-VL~\citep{qwenvl} & Qwen-7B &ViT-G&\xmark & -- & -- & 78.8 & -- & 59.3 & 35.2 &  -- & 32.9 & 13.0  \\

Qwen-VL-Chat~\citep{qwenvl} & Qwen-7B &ViT-G&\cmark & -- & -- & 78.2 & -- & 57.5& 38.9  & -- & 59.1 & --  \\

\midrule[0.6pt]
\multicolumn{13}{c}{\textit{Models for both Image and Text Generation}}\\ 
\midrule[0.6pt]
CM3Leon~\citep{cm3leon} & -- & -- &\cmark &
61.6  & 10.5 &
47.6 & 23.8 & -- & 37.6 &  22.6 & -- & --   \\

Emu~\citep{emu} & Vicuna-13B &ViT-L&\cmark &
112.4  & -- & 
52.0 & 38.2 & -- & 34.2 &  47.4 & -- & --  \\

Emu-I~\citep{emu} & Vicuna-13B &ViT-L&\cmark &
\textbf{117.7}  & -- & 
40.0 & 34.7 & -- & 35.4 &  48.0 & -- & -- \\

Emu2~\citep{emu2} & LLaMA-33B &ViT-L&\cmark &
--  & -- & 
33.3 & 26.7 & -- & 40.4 &  -- & -- & -- \\


DreamLLM~\citep{dreamllm} & 
Vicuna-7B &ViT-L&\xmark & 103.7 & 8.4 & \textbf{72.9} & 52.2
& -- & 49.3 &  -- & 58.2 & 36.6 \\

\midrule[0.6pt]
\name-VQA & Vicuna-7B & ConvNext-B & \xmark & 
115.4 & \textbf{22.4} & 68.2 & \textbf{53.4} & \textbf{55.7} & \textbf{50.4} & 42.1 & \textbf{60.8} & \textbf{37.4} \\ 
\bottomrule 
\end{tabular}}
\vspace{-0.25cm}
\label{tab:multi_tasks}
\end{table*}

\textbf{Image-Level Visual Question Answering and Captioning.} In order to assess multimodal vision and language capabilities of \name, we conduct evaluation against current SOTA LMMs including LLaVA-1.5~\citep{llava2}, Qwen-VL~\citep{qwenvl}, DreamLLM~\citep{dreamllm} and MM-Interleaved~\citep{mm-interleaved} across several tasks, including image captioning on COCO ~\citep{coco}, Image2Paragraph~\citep{img2paragraph}, visual question answering on VQAv2~\citep{vqav2}, OKVQA~\citep{ok-vqa}, GQA~\citep{gqa}, VizWiz~\citep{vizwiz}, and VisDial~
\citep{visdial}. As demonstrated in \cref{tab:multi_tasks}, \name exhibits
superior or comparable performance relative to SOTA models. In comparison with models for text generation only, our approach consistently achieves competitive performance across various dataset splits. Against models for both image and text generation, \name demonstrates enhanced performance in nine dataset splits. Compared to the current state-of-the-art model DreamLLM, DEEM outperforms DreamLLM in six out of the seven shared evaluation dataset splits. It is noteworthy that \name is trained with a significantly smaller image encoder CLIP-ConvNeXt-B~\citep{convnext}, comprising only 122M parameters, in stark contrast to baselines such as DreamLLM~\citep{dreamllm}, which utilize larger 427M CLIP-ViT-L~\citep{clip}. These results indicate that our method can enhance the model's robustness performance without compromising the multimodal vision and language capabilities of our model.

\subsection{Ablation Study}
\vspace{-4pt}
In this study, we conduct ablation studies on several key components of the model, including consistency semantic regularization, training latency, scalability and the impact of different architectures. Benchmarks include RobustVQA-A:RVQA-A; RobustVQA-R: RVQA-R; RobustVQA-V:RVQA-V; POPE-R~\citep{pope}; POPE-P~\citep{pope}; POPE-A~\citep{pope}; MMVP~\citep{mmvp}; OK-VQA~\citep{ok-vqa}. More additional ablation studies can be found in \cref{app:add_ablation_study}.

\begin{table*}[!b]
\centering
\footnotesize
\setlength\tabcolsep{2pt}
\caption{\textbf{Ablation study of $\mathcal{L}_{\text{CSR}}$ and training latency. } Using semantic consistency regularization during both the pre-training and supervised fine-tuning phases can significantly enhance the model's robustness and resistance to hallucinations, while incurring only a marginal additional training cost.}\label{tab:ablation_csr_latency}
\resizebox{0.95\linewidth}{!}{
    \begin{tabular}{ll|ccccccc}
    \toprule
    SFT & $\mathcal{L}_{\text{CSR}}$ & RVQA-A & RVQA-R & RVQA-V & POPE-R & POPE-P & POPE-A & SPEED \\
    \midrule[0.5pt]
    \xmark & \xmark & 53.2 & 56.1 & 54.7 & 50.6 & 52.0 & 51.9 & \textbf{8.11} s/step\\
    \xmark & \cmark & \textbf{57.8} & \textbf{69.0} & \textbf{64.8} & \textbf{69.9} & \textbf{70.3} & \textbf{68.9} & 9.25 s/step \\
    \midrule[0.5pt]
    \cmark & \xmark & 51.3 & 56.5 & 57.4 & 85.4 & 78.8 & 76.2 & \textbf{2.14} s/step \\
    \cmark & \cmark & \textbf{53.5} & \textbf{57.6} & \textbf{58.1} & \textbf{86.0} & \textbf{79.2} & \textbf{77.1} & 2.22 s/step\\
    \bottomrule
    \end{tabular}
    }
\vspace{-15pt}
\end{table*}

\textbf{Consistency Semantic Regularization and Training Latency.} To evaluate the effectiveness of the key elements of our design, we conduct the following ablation experiments. We first pre-train a baseline model without using the consistency semantic regularization term under the same training setting for comparison to demonstrate the effectiveness of our architecture. As we can see from \cref{tab:ablation_csr_latency}, during the pre-training phase, using our consistency semantic regularization can significantly enhance the model's performance on both hallucination and robustness benchmarks. Moreover, we load the weights of the pre-trained model for image-text instruction fine-tuning experiments. In the second phase of image-text instruction fine-tuning experiments, we demonstrate the effectiveness of our model design. As detailed in \cref{tab:ablation_csr_latency}, we observe that after fine-tuning with image-text instruction data, the model's visual hallucination ability improves further, but its visual perception robustness decreases. However, using our consistency semantic regularization can mitigate the robustness degradation while further enhancing the model's visual hallucination ability. To explore the impact of introducing consistency semantic regularization on the training latency in the two stages of training, we conduct corresponding ablation experiments. We present the result in ~\cref{tab:ablation_csr_latency}. Employing consistency semantic regularization adds only a marginal increase in training latency, yet it significantly enhances the model's robustness capabilities.

\begin{table*}[!h]
\centering
\footnotesize
\setlength\tabcolsep{2pt}
\caption{\textbf{Ablation study of model scalability. } Gradually expanding the training data and model size can further enhance the model's capabilities, demonstrating the scalability of the approach.}\label{tab:scalability}
\vspace{-6pt}
\resizebox{1.0\linewidth}{!}{
\begin{tabular}{ll|ccccccc}
    \toprule
    Architecture & Training Data & RVQA-A & RVQA-R & RVQA-V & POPE-R & POPE-P & POPE-A & OK-VQA \\
    \midrule[0.5pt]
    ConvNext-B/Vicuna 7B & 32K & 51.86 & 54.31 & 52.73 & 48.44 & 50.10 & 50.06 & 20.74\\
    ConvNext-B/Vicuna 7B & 96K & 52.31 & 57.43 & 54.06 & 54.42 & 57.22 & 56.35 & 22.33 \\
    ConvNext-B/Vicuna 7B & 160K & 52.89 & 58.93 & 55.31 & 60.28 & 60.74 & 59.96 & 23.65 \\
    ConvNext-L/Vicuna 7B& 160K & 53.23 & 60.47 & 56.88 & 61.12 & 62.87 & 62.09 & 23.87\\
    ConvNext-B/Vicuna 13B& 160K & \textbf{53.92} & \textbf{61.27} & \textbf{57.02} & \textbf{62.60} & \textbf{64.26} & \textbf{63.19} & \textbf{31.11}\\
    \bottomrule
    \end{tabular}}
\vspace{-12pt}
\end{table*}

\textbf{Model Scalability.} Although DEEM demonstrates better performance with smaller data count and model sizes, its scalability has yet to be validated. As is well known, scalability is crucial for model performance. We conduct ablation experiments to assess the scalability concerning data count and model size. As shown in \cref{tab:scalability}, gradually increasing the training data enables the model to successfully scale while achieving improved results. Additionally, increasing the sizes of both the VFM and LLM leads to sustained performance enhancements, indicating that DEEM possesses good scalability.

\begin{table*}[!h]
\centering
\footnotesize
\setlength\tabcolsep{2pt}
\caption{\textbf{Ablation study of different architectures.} Our method not only significantly enhances the capabilities of LLMs for text and image generation with marginal additional training costs, but it also improves the performance of LLMs for text generation only, validating the generalization ability of the approach.}\label{tab:generalization}
\resizebox{0.95\linewidth}{!}{
    \begin{tabular}{ll|ccccccc}
    \toprule
    Name & $\mathcal{L}_{\text{CSR}}$ & MMVP & RVQA-A & RVQA-R & RVQA-V & POPE-R & POPE-P & POPE-A \\
    \midrule[0.5pt]
    LLaVA & \xmark & 18.6 & 54.8 & 60.0 & 58.7 & 55.5 & 53.3 & 54.6 \\
    LLaVA & \cmark & \textbf{25.1}  & \textbf{56.7} &  \textbf{66.7} & \textbf{61.9} & \textbf{67.9} & \textbf{68.7} & \textbf{65.4} \\
    \midrule[0.5pt]
    DEEM & \xmark & 20.3  & 53.2 & 56.1 & 54.7 & 50.6 & 52.0 & 51.9 \\
    DEEM & \cmark & \textbf{28.7}  & \textbf{56.9} & \textbf{68.6} & \textbf{63.1} & \textbf{69.9} & \textbf{70.3} & \textbf{68.9} \\
    \bottomrule
    \end{tabular}}
\vspace{-5pt}
\end{table*}

\textbf{Impact of Different Architectures.} By cleverly reusing the diffusion model from LMMs for image and text generation, we can significantly enhance the model's foundational visual perception, visual robustness, and anti-hallucination capabilities with only marginal additional training costs. However, whether DEEM possesses sufficient generalization ability to remain effective for LMMs on text generation only has yet to be explored. To validate our hypothesis, we employ the LLaVA~\citep{llava} architecture and conducted ablation experiments using semantic consistency regularization loss, with results presented in \cref{tab:generalization}. We observe that utilizing diffusion feedback to improve the basic perceptual capabilities of LMMs—thus preventing the model from overly compressing visual information and losing sensitivity to subtle details—is a general method that is architecture-agnostic and exhibits good generalization properties. This suggests that the benefits of our approach could extend beyond the specific configurations tested, potentially enhancing a wide range of LMMs in various applications.

\vspace{-6pt}
\section{Related Work}\label{sec:related}
\vspace{-6pt}
\subsection{Diffusion Models for Representation Learning}
Diffusion models have made significant progress in various generative tasks~\citep{song2020score,ho2020denoising}, such as image generation~\citep{betker2023improving}, video generation~\citep{ho2022imagen}, and object tracking~\citep{diffusiontrack}. In addition to the aforementioned research, many studies focus on leveraging diffusion models for representation learning. Some works utilize the conditional control of pre-trained diffusion models to flexibly address different downstream tasks, including object classification~\citep{xiang2023denoising}, semantic segmentation~\citep{xu2023open}, image caption~\citep{wei2024diffusion}, and keypoint matching~\citep{nam2023diffusion}. Other studies ~\citep{li2023self, song2024autogenic} design specialized modules and train diffusion models from scratch to further enhance representation capabilities. Although diffusion models have been widely applied in the generative tasks of  large multimodal models, the use of diffusion models to optimize the visual representations of large multimodal models has yet to be explored. To our knowledge, we are the first to employ diffusion models in a self-supervised paradigm to optimize the visual representations of  large multimodal models, significantly enhancing their perceptual abilities and reliability at minimal cost.

\subsection{Large Multimodal Model}

Image-to-text large multimodal models (LMMs)~\citep{luo2025openomni,liu2024iibench,zhang2023ideal,wang2024open,zhou2024few,liu2024towards} inject visual information into large language models (LLMs) through vision foundation models (VFMs), allowing the language models to perceive visual inputs and thus generate captions or answer questions based on the given multimodal content. Flamingo~\citep{flamingo} tries to extract vision features with a resampler, and transfer them into the text features with a cross-attention mechanism. Instead of using cross-attention layers, BLIP-2~\citep{blip2} directly feed the visual features into the LLMs as soft prompts and significantly reduce the training cost by reducing the visual token number. LLaVA~\citep{llava} and MiniGPT-4~\citep{minigpt4} construct a small-scale instruction tuning dataset to better align the LMM with the expected output format. 
Although this unidirectional image-to-text paradigm has achieved tremendous success, it still fails to unify multimodal tasks like text-to-image generation and image-to-text visual question answering, significantly limiting the capabilities of multimodal models.

In order to unify multimodal tasks into a unified manner, some works~\citep{cm3leon,GILL,emu,dreamllm,mm-interleaved,seed,seed-x,mmevol} attempt to generate images and text in the interleaved context concurrently. The release of some public large-scale interleaved image-text datasets~\citep{obelics,mmc4} has significantly advanced the development of this field. CM3Leon~\citep{cm3leon} converts images into discrete tokens, facilitating token-level auto-regressive modeling as traditional language modeling. Although CM3Leon showcases competitive image generation capabilities, it exhibits notable weaknesses in image understanding. Emu~\citep{emu} and DreamLLM~\citep{dreamllm} focus on single-stage end-to-end modeling using raw image pixels as input for interleaved image-text generation modeling, but they feed image information at the input of LMMs, which are limited by the problem that fixed number of visual tokens cannot efficiently describe image details. MM-Interleaved~\citep{mm-interleaved} addresses this limitation by integrating image details into LMMs via multi-scale visual features. However, when faced with out-of-distribution noisy data, the image encoders used by LMMs often produce incorrect visual information, ultimately leading to erroneous predictions. This significantly limits the application of the models in safety-critical scenarios. Building on an advanced interleaved content modeling mechanism, we propose \name, which cleverly reuses DMs to correct the outputs of the VFMs without increasing extra parameter count, thereby enhancing the model's generalization capabilities and reducing visual hallucinations in a self-supervised manner. Similar to previous work~\citep{llava,dreamllm,mm-interleaved}, after supervised fine-tuning, it achieves competitive performance on multiple downstream multimodal tasks with the smallest scale.

\vspace{-6pt}
\section{Conclusion}
\vspace{-6pt}
Can diffusion models serve as the eyes of
large language models for image perception? In this paper, we answer the question by proposing a novel method called \name, which leverages a diffusion model as the eyes for LLMs. This approach enhances the robustness of the multimodal model for interleaved image-text modeling and reduces visual hallucinations without introducing extra modules. Through comprehensive exploratory experiments, we demonstrate the effectiveness of
the proposed \name method. In addition to its advanced robust performance and visual hallucination handling capabilities, we adopt an additional two-stage instruction fine-tuning process to broaden the application scenarios of our \name. This enables \name to handle a variety of multimodal tasks, including visual question answering, image captioning, and region-level image reasoning. Besides, this work initiates the first step towards visual robustness via generative feedback in a multimodal model. In the future, we will continue to enhance the model’s ability to conduct better multimodal comprehension and creation tasks. As an end-to-end framework, we hope it will spur further research in the multimodal robustness field, such as multimodal agents that can handle complex tasks that require safety abilities. 


\section{Acknowledgments}
Min Yang is supported by National Key Research and Development Program of China (2022YFF0902100), National Natural Science Foundation of China (Grant No. 62376262), the Natural Science Foundation of Guangdong Province of China (2024A1515030166).
Xiaobo Xia is supported by MoE Key Laboratory of Brain-inspired Intelligent Perception and Cognition, University of Science and Technology of China (Grant No. 2421002). 

\bibliography{iclr2025_conference}
\bibliographystyle{iclr2025_conference}
\newpage
\tableofcontents

\appendix

\section{Limitation}\label{app:limitation}
Although our method significantly enhances the visual robustness of interleaved image-text modeling multimodal models after image-text alignment pre-training, it, unfortunately, cannot eliminate but only alleviate the robustness knowledge forgetting issue caused by subsequent fine-tuning, as shown in the \cref{tab:ablation_csr_latency}. Additionally, our model requires using a diffusion model as another eye to correct and update the erroneous knowledge of the image encoder to improve the overall visual robustness of the multimodal model. However, updating larger image encoders such as CLIP-ViT-L and CLIP-ViT-G\citep{clip} will increase the training overhead, which may limit the application of our model. We hope that in the future, the diffusion model can completely replace the image encoder to further enhance the effectiveness of our method.

\section{Broader Impacts}\label{impacts}
The proposed method introduces a novel strategy to enhance the robustness and generalization capabilities of multimodal models by leveraging a diffusion model as an additional eye for large language models. This strategy allows for the correction and updating of potential semantic errors in the image encoder, leading to significant improvements in handling out-of-distribution data and mitigating visual hallucinations. Overall, our contributions provide a significant step forward in the field of multimodal, offering a robust, efficient, and scalable solution for improving the accuracy and reliability of multimodal models. The broader impacts of this work include the potential for more intelligent and adaptive AI systems that can operate effectively in diverse and challenging environments.

\begin{table*}[!h]
\centering
\footnotesize
\setlength\tabcolsep{2pt}
\caption{\textbf{Zero-shot region-level image captioning results} on ReferCOCOg.}\label{tab:region_cap}
\resizebox{0.5\linewidth}{!}{
    \begin{tabular}{lccc}
    \toprule[0.95pt]
    Method & Type & METEOR & CIDEr \\
    \midrule[0.6pt]
    GRIT~\citep{grit} &  Box & 15.2 & 71.6 \\
     Kosmos-2 (0-shot)~\citep{kosmos2} & Box & 12.2 & 60.3 \\
     Kosmos-2 (2-shot)~\citep{kosmos2} &  Box & 13.8 & 62.2 \\
    Kosmos-2 (4-shot)~\citep{kosmos2} &  Box & 14.1 & 62.3 \\
    NeXt-Chat~\citep{nextchat} & Box & 12.0 & 79.0 \\
    \midrule[0.6pt]
    \name-Mask & Mask & 14.1 & 71.0  \\
    \bottomrule[0.95pt]
    \end{tabular}
    }
\vspace{-20pt}
\end{table*}

\section{Additional Experiments Results}\label{app:add_exp}


\begin{wrapfigure}{r}{0.42\textwidth}
    \vspace{-1.2cm}
    \centering
    \setlength\tabcolsep{3pt}
    \makeatletter\def\@captype{table}\makeatother
    \caption{\textbf{Zero-shot text-to-image generation FID} on MS-COCO and LN-COCO.
    }\label{tab:coco_text2img}
    \vspace{-0.10cm}
    \resizebox{\linewidth}{!}{
    \begin{tabular}{lcc}
    \toprule[0.95pt]
    Method & MS-COCO & LN-COCO\\
    \midrule[0.6pt]
    \multicolumn{3}{c}{\textit{Text-to-Image Specialists}}\\
    \midrule[0.6pt]
    Retrieval Result & 17.97 & 33.59 \\
    DALL-E~\citep{dalle} & $\sim$28 & - \\
    CogView~\citep{cogview}  & 27.10 & - \\
    CogView2~\citep{cogview2} & 24.00 & - \\
    Stable Diffusion~\citep{stablediffusion} & 12.43 & 34.26\\
    GLIDE~\citep{glide} & 12.24 & -\\
    Make-A-Scene~\citep{make-a-scene} & 11.84 & -\\
    DALL-E 2~\citep{dalle2} & 10.39 & -\\
    Muse-3B~\citep{muse} & 7.88 & -\\
    Imagen-3.4B~\citep{imagen} & 7.27 & -\\
    Parti-20B~\citep{parti} & 7.23 & 15.97 \\
    \midrule[0.6pt]
    \multicolumn{3}{c}{\textit{Models for both Image and Text Generation}}\\
    \midrule[0.6pt]
    CM3-13B~\citep{cm3} &  29.56 & -\\
    GILL-8B~\citep{GILL} & 12.25 & -\\
    Emu-13B~\citep{emu} & 11.66& -\\
    CM3Leon-7B~\citep{cm3leon} & 10.82 & -\\
    DreamLLM-7B~\citep{dreamllm} & 8.76 & 22.42\\
    \midrule[0.6pt]
    \name-7B (Ours) & 8.89 & 24.13\\
    \bottomrule[0.95pt]
    \end{tabular}
}
\vspace{-0.55cm}
\end{wrapfigure}

\textbf{Region-Level Image Captioning.} In addition to holistic image understanding, we also validate the model’s ability to take region-level image captioning. As shown in \cref{fig:mask_aware_extractor}, we use a mask-aware extractor to obtain region-level visual features and address region-level image captioning tasks. We adopt the RefCOCOg~\citep{refercoco+g} validation set and compare it with other state-of-the-art (SOTA) models, including GRIT~\citep{grit}, Kosmos-2~\citep{kosmos2}, and NeXt-Chat~\citep{nextchat}. The CIDEr~\citep{cider} and METEOR are applied as the evaluation metrics. As shown in \cref{tab:region_cap}, our model is capable of achieving competitive performance on CIDEr and METEOR across all of the compared methods, which shows the superiority of our \name.

\textbf{Text-to-Image Generation}. we evaluate text-conditional image generation on MS-COCO~\citep{mscoco} and LN-COCO~\citep{lncoco}. On MSCOCO, we sample 8 images per text condition and use CLIP-ViT-L~\citep{clip} to rerank based on text-image similarity. CLIP reranking is not used for LN-COCO. FID ~\citep{fid} is used to evaluate both datasets. As shown in \cref{tab:coco_text2img}, our model shows
competitive text-to-image generation compared to existing
image and text generation models. See
qualitative results on text-to-image synthesis in \cref{fig:text_to_image_case} in \cref{app:add_vis_example}.

\section{Additional Ablation Study}\label{app:add_ablation_study}
we provide more ablation studies for \name in this section, all of which share the same settings. All the code, models, and data tools will be released soon.

\subsection{Ablation Study of Input Image Resolution}

\begin{table*}[!t]
\centering
\footnotesize
\setlength\tabcolsep{2pt}
\caption{Ablation study of input image resolution 
 and coefficient $\lambda$ with 2k training steps and 16 batch size.}\label{tab:ablation_resolution_coeffient}
\resizebox{0.95\linewidth}{!}{
    \begin{tabular}{lcc|ccccccc}
    \toprule
    SFT & resolution & $\lambda$ & RVQA-A & RVQA-R & RVQA-V & POPE-R & POPE-P & POPE-A & OK-VQA\\
    \midrule[0.5pt]
    \xmark & 256 & 1 & 51.6 & 52.0 & 49.6 & 48.5 & 50.0 & 50.0 & 18.9\\
    \xmark & 256 & 5 & \textbf{51.9} & \textbf{54.3} & \textbf{52.7} & 48.4 & 50.1 & 50.0 & \textbf{20.7}\\
    \xmark & 256 & 10 & 51.7 & 52.7 & 51.9 & \textbf{48.7} & \textbf{50.3} & \textbf{50.3} & 20.1\\
    \midrule[0.5pt]
    \cmark & 256 & 5 & 51.5 & \textbf{59.1} & 57.9 & 85.9 & 77.1 & 76.4 & 38.7\\
    \cmark & 448 & 5 & \textbf{52.5} & 57.6 & \textbf{58.1} & \textbf{86.0} & \textbf{79.2} & \textbf{77.1} & \textbf{41.0}\\
    \bottomrule
    \end{tabular}}
\vspace{-20pt}
\end{table*}


In addition to the aforementioned exploration, we also scale up the input image resolution for performance gain. The performance gain becomes larger when further increasing the input image resolution from 256 to 448 in image-text instruction fine-tuning, as shown in \cref{tab:ablation_resolution_coeffient}. Such results indicate our method could better exploit the additional information gained from high resolution. Moreover, we conduct an ablation study on coefficient $\lambda$ in loss function. As shown in \cref{tab:ablation_resolution_coeffient}, setting $\lambda$ = 5
achieves a better balance between robustness and hallucination empirically.

\subsection{Ablation Study of Training Recipes}
We also conduct an ablation study to control the trainability of different training modules. As shown in \cref{tab:ablation_training_recipe}, we found that freezing the DM (Diffusion Model) while not freezing the VFM (Visual Foundation Model) during training yields the best robustness and hallucination results.
\section{Additional Implementation Details}\label{app:imple_details}


\begin{table*}[b!]
    \centering
    \setlength\tabcolsep{5pt}
    \caption{
    \textbf{Comparison of different VQA formats.} Questions in the yes or no format can well evaluate the performance of the models on the RobustVQA benchmark, while questions in the multiple-choice format are very random, and MM-interleaved tend to output the first option. Therefore, we adopt yes or no format in our experimental settings. More details about the new benchmark RobustVQA design can be found in \cref{subsec:dataset_construction}.}\label{tab:data_format}
    \vspace{-1pt}
    \resizebox{\linewidth}{!}{
    \begin{tabular}{llccc}
    \toprule[0.95pt]
    Format & Prompt & RobustVQA-A & RobustVQA-R & RobustVQA-V\\
    \midrule[0.6pt]
     \multirow{2}{*}{multiple-choice} & ``\texttt{What is the main object in this image?}'' & \multirow{2}{*}{44.88} & \multirow{2}{*}{58.88} & \multirow{2}{*}{46.86} \\[6pt]
     & ``\texttt{Chose from the list: [false category label,gt category label].}'' &  &  & \\[6pt]
     \multirow{2}{*}{multiple-choice} & ``\texttt{What is the main object in this image?}'' & \multirow{2}{*}{\textbf{84.60}} & \multirow{2}{*}{\textbf{90.16}} & \multirow{2}{*}{\textbf{82.92}} \\[6pt]
     & ``\texttt{Chose from the list: [gt category label,false category label].}'' &  &  & \\[6pt]
     \multirow{4}{*}{yes or no} & ``\texttt{Is [gt category label] the main object in this image?}'' & \multirow{4}{*}{50.76} & \multirow{4}{*}{52.71} & \multirow{4}{*}{50.60} \\[6pt]
     & ``\texttt{Please answer yes or no.}'' &  &  & \\[6pt]
    & ``\texttt{Is [false category label] the main object in this image?}'' &  &  & \\[6pt]
  & ``\texttt{Please answer yes or no.}'' &  &  & \\[6pt]
    \bottomrule[0.95pt]
    \end{tabular}}
\vspace{-15pt}
\end{table*}

\subsection{Dataset Construction}\label{subsec:dataset_construction}

\begin{figure*}[h]
    \centering
    \includegraphics[width=\linewidth]{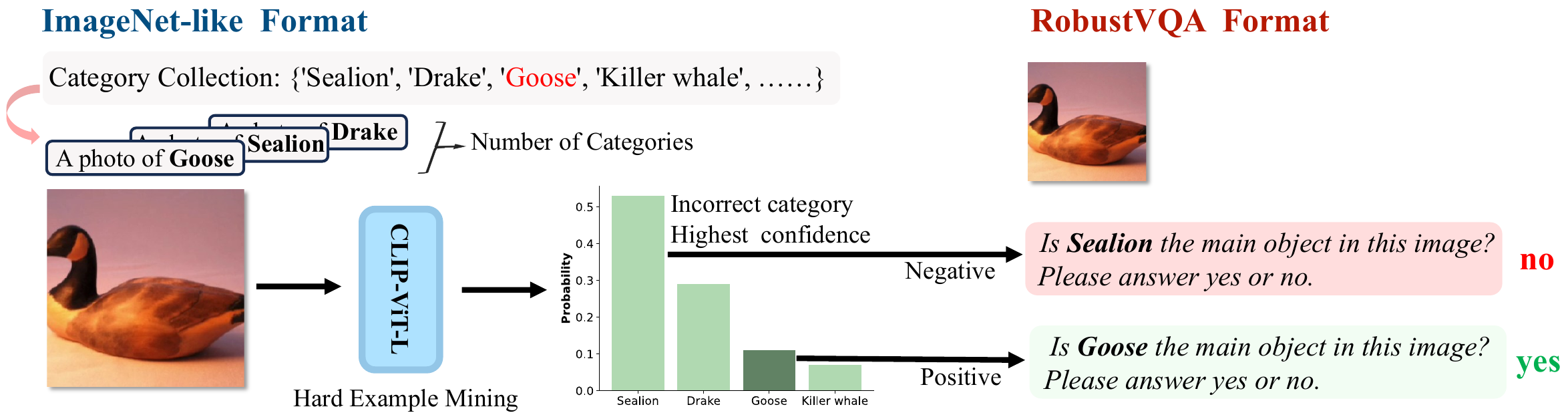}
    \vspace{-0.55cm}
    \caption{\textbf{Robustness dataset construction process.} We use the CLIP-ViT-L model for hard example mining and then transform them into question-answer pairs via a pre-defined template.}
\label{fig:dataset_construction}
\vspace{-0.35cm}
\end{figure*}

As shown in \cref{fig:dataset_construction}, we first convert the original ImageNet-A~\citep{imagenet-a}, ImageNet-R~\citep{imagenet-r}, and ImageNet-V2~\citep{imagenet-v2} data into a VQA format that the multimodal model can evaluate. Specifically, we use the CLIP-ViT-L model for hard example mining, predicting the incorrect category label with the highest confidence score apart from the ground truth category label. We then use a pre-defined prompt as: ``\texttt{Is [category label] the main object in this image? Please answer yes or no.}'' to simultaneously construct a pair of positive and negative example samples, allowing the model to answer ``\texttt{yes}" or ``\texttt{no}". By using this design, we can evaluate the robustness of multimodal models in an unbiased manner with the new benchmark called RobustVQA, facilitating both assessment and comparison. It is worth noting that, as shown in \cref{tab:data_format}, we find that the yes or no format is more stable than the multiple-choice format and can better evaluate the robustness of multi-modal models. 

\subsection{Image-Text Alignment Pre-training}
\begin{wrapfigure}{r}{0.56\textwidth}
    \vspace{-0.45cm}
    \centering
    \setlength\tabcolsep{3pt}
    \makeatletter\def\@captype{table}\makeatother
    \caption{Ablation study of training recipe in image-text alignment pre-training with 10k training steps and 128 batch size.}\label{tab:ablation_training_recipe}
    \vspace{-0.14cm}
    \resizebox{\linewidth}{!}{
    \begin{tabular}{ll|cccccc}
    \toprule
    VFM& DM & RVQA-A & RVQA-R & RVQA-V & POPE-R & POPE-P & POPE-A\\
    \midrule[0.5pt]
    freeze & unfreeze & 53.2 & 56.1 & 54.7 & 50.6 & 52.0 & 51.9\\
    unfreeze & freeze & \textbf{56.8} & \textbf{68.6} & \textbf{63.1} & \textbf{69.9} & \textbf{70.3} & \textbf{68.9}\\
    unfreeze & unfreeze & 50.3 & 52.5 & 53.1 & 54.8 & 56.3 & 56.1\\
    \bottomrule
    \end{tabular}
}
\vspace{-0.45cm}
\end{wrapfigure}
We use MMC4-Core~\citep{mmc4}, LAION-400M~\citep{laion-400m}, SBU~\citep{sbu}, and CC-12M~\citep{cc} as
the pre-training dataset. For LAION-400M~\citep{laion-400m}, SBU~\citep{sbu}, and CC-12M~\citep{cc}, instead of utilizing the original annotations, we use the version filtered by the pre-trained BLIP-2 model~\citep{blip2}. For simplicity, we refer to it as BLIP-LCS hereafter. "LCS" abbreviates the LAION, CC, and SBU datasets. Text prompts with lengths shorter
than 10 are also filtered out. Due to network constraints, we only collect approximately 6M of MMC4-Core and 20M of BLIP-LCS data. The sampling probability of MMC4 is twice that of BLIP-LCS. The images are inserted before or after the corresponding text sentence with equal probability. Specifically, images with a CLIP similarity score below 0.24 will be discarded, and only 6 images at most will be kept for each document in MMC4-Core. We also exclude 100\% of all documents that do not contain any images, and 50\% of documents that contain only 1 image. For image-text-pair BLIP-LCS
datasets, we randomly sample multiple image-text pairs from the same
dataset and concatenate them to the maximum context
length (\ie, 2048) during pre-training. For interleaved image and text MMC4-Core~\citep{mmc4} datasets, we also split and concatenate the documents to form the training samples. Such a concatenation strategy can utilize the full context window of Large Language Models and thus achieve high data efficiency. Besides that, for image
generation, we ignore the training loss of images which are
the first element in the sequence. The text condition of the
rest images is dropped with a 10\% probability to improve
classifier-free guidance sampling. The detailed hyper-parameters of image-text alignment pre-training are listed in \cref{tab:details}.

\subsection{Image-Text Instruction Fine-tuning}
We utilize public available datasets for supervised fine-tuning, including LLaVA-665K\citep{llava}, COCO Caption~\citep{coco}, VQAv2~\citep{vqav2},TextCaps~\citep{textcaps}, OCR-VQA~\citep{ocrvqa}, GQA~\citep{gqa}, OK-VQA~\citep{ok-vqa}, TextVQA~\citep{textvqa}, and AOK-VQA~\citep{aokvqa}. We use the following prompt template \texttt{``Based on the image, please answer the
question. \{image\} \{question\}. The answer is: \{answer\}
"} to convert the data into a mixture of instruction following forms, resulting in approximately 800K instruction data for the second-stage image-text instruction fine-tuning. The detailed hyper-parameters of
image-text instruction fine-tuning are listed in \cref{tab:details}.

\subsection{Mask-Text Instruction Fine-tuning}
 We collect short text and pixel-level mask pairs from the publicly available object-level datasets (COCO, RefCOCO,
RefCOCO+) and part-level datasets (Pascal Part, Part Imagenet), then transform them into instruction following data. Moreover,  Visual Genome (VG) and Visual Commonsense
Reasoning (VCR) datasets are employed to add more multiple region understanding data, resulting in approximately 200K instruction data for the third-stage mask-text instruction fine-tuning. See more hyper-parameters details in \cref{tab:details}.
\subsection{Evaluation}

\begin{figure*}[h]
    \centering
    \includegraphics[width=\linewidth]{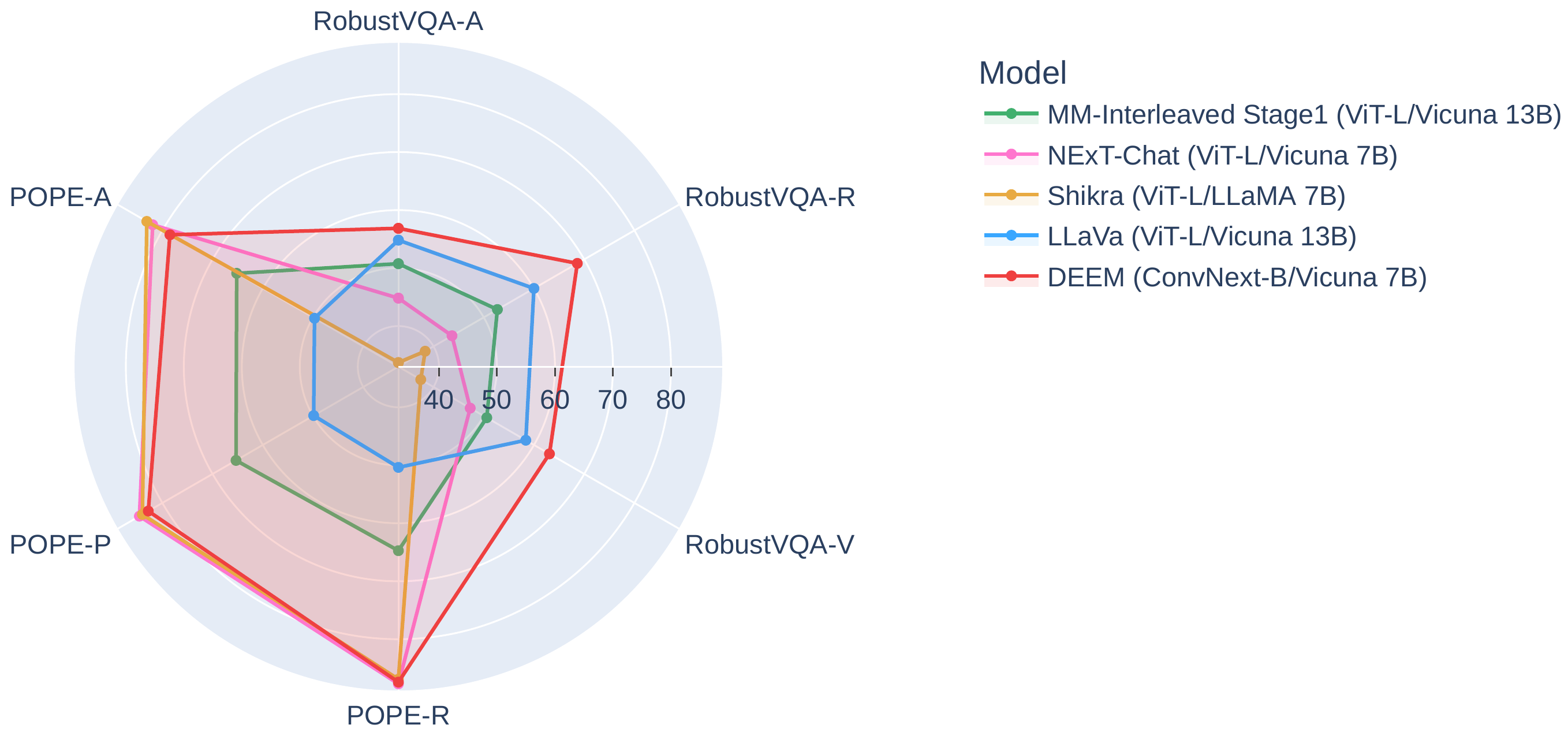}
    \vspace{-0.55cm}
    \caption{\textbf{Performance on visual robustness and hallucination benchmark.} \name achieves the best results on robustness benchmark and competitive performance on hallucination even at the smallest scale, demonstrating the efficiency and effectiveness of our approach.}
\label{fig:radar_exp}
\vspace{-0.35cm}
\end{figure*}

\begin{figure*}[h]
    \centering
    \includegraphics[width=\linewidth]{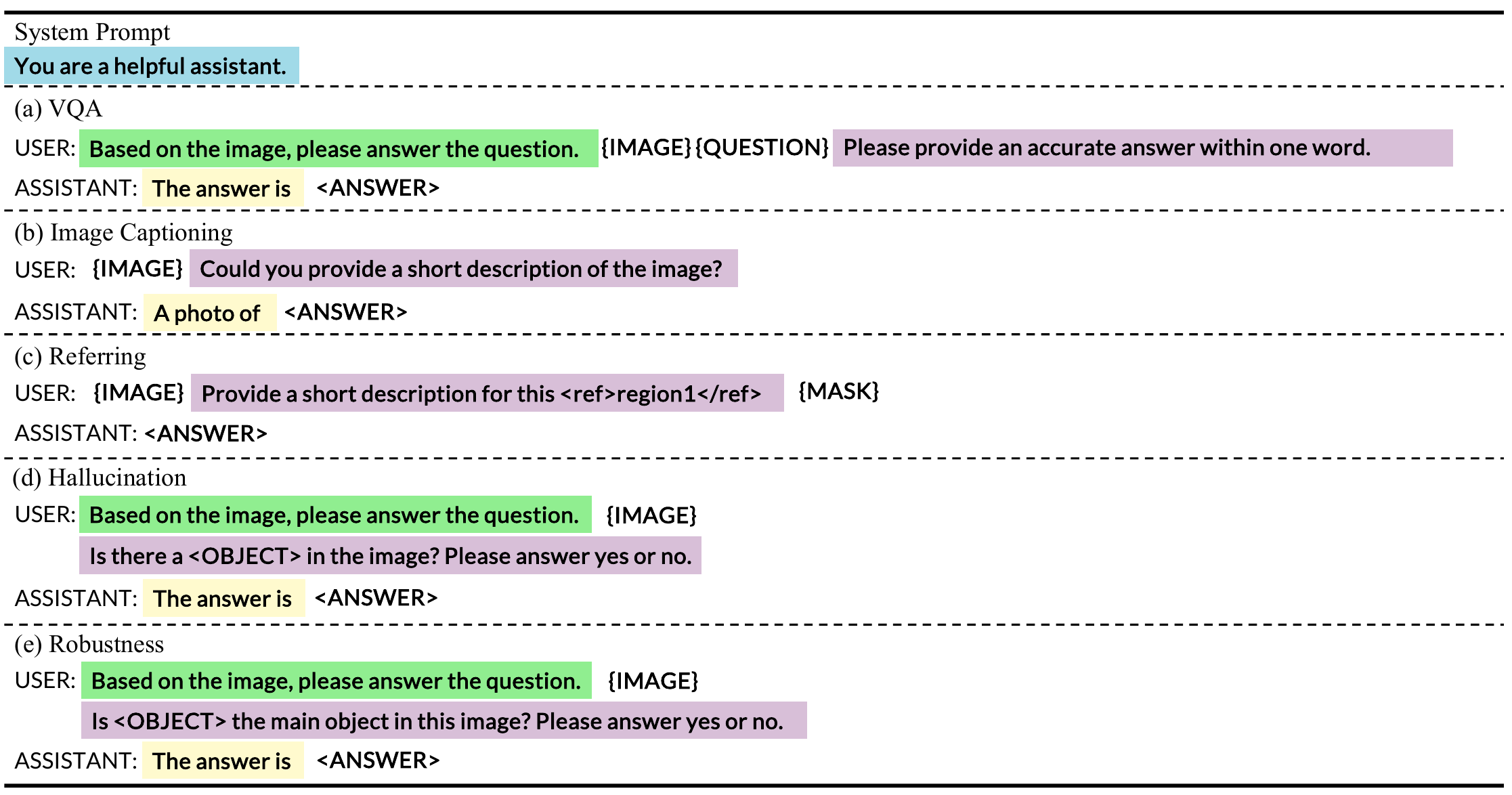}
    \caption{\textbf{Prompt template used for evaluation.}  (a) VQA includes VQAv2, VizWiz, OKVQA, GQA, VisDial, and MMVP. (b) Image Captioning includes COCO, Image2Paragraph. (c) Region-level Image Captioning includes RefCOCOg. (d) Visual hallucination includes POPE. (e) Visual Robustness includes RobustVQA-A, RobustVQA-R, and RobustVQA-V. $<\text{IMAGE}>$denotes the input image representation, $<\text{MASK}>$ denotes the mask-level image representation, $<\text{QUESTION}>$denotes each specific question, $<\text{ANSWER}>$ is the generated answer, and 
    $<\text{OBJECT}>$ is the specific object name in a question of POPE and RobustVQA.}
\label{fig:template}
\end{figure*}

As shown in \cref{fig:radar_exp}, \name achieves the
best results on both hallucination and robustness benchmarks even at the smallest scale, demonstrating
the efficiency and effectiveness of our approach. In addition to visual robustness and hallucination, we also use various benchmarks and datasets, such as
image caption, visual question answering, text-to-image
generation and so on, to assess the image-text comprehension capabilities. All these evaluation tasks and metrics are listed in \cref{tab:eval_benchmarks_summary}. The prompt templates for each task are listed in \cref{fig:template}.

\begin{figure*}[h]
    \centering
    \includegraphics[width=\linewidth]{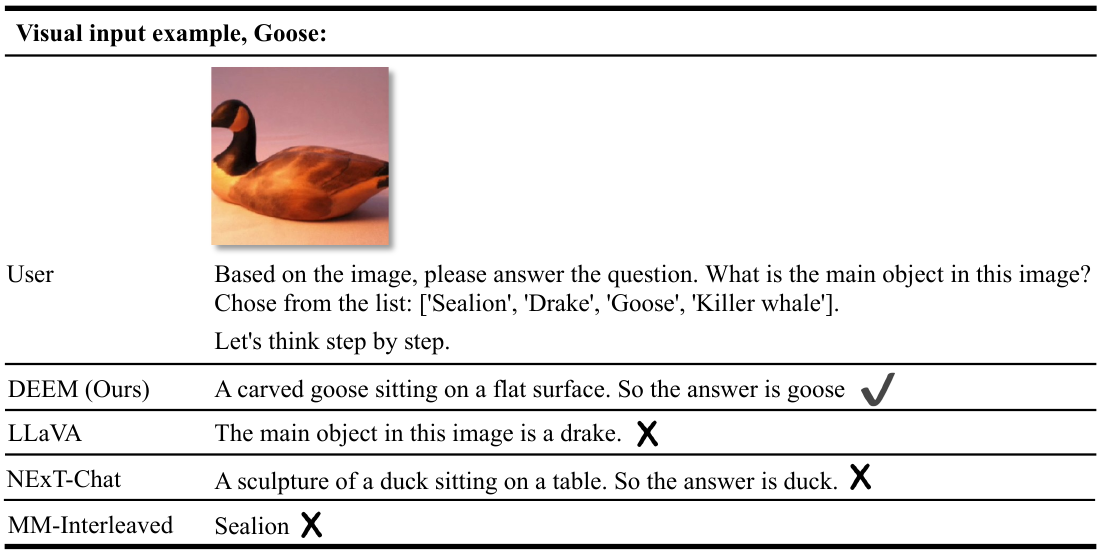}
    \caption{\textbf{Case Comparison.} Compared to other SOTA models, including LLaVA, NeXt-Chat, and MM-Interleaved, when encountering out-of-distribution data, their models are affected by incorrect semantics from the image encoder and cannot output the correct answer. However, \name can output the correct answer via generative feedback.}
\label{fig:case_comparison}
\end{figure*}

\begin{table*}[t!]
\caption{\textbf{Training recipes} for \name. The three training stages are introduced in \cref{subsec:training}. \texttt{Stage I}: Image-Text Alignment Pre-training, \texttt{Stage II}: Image-Text Instruction Fine-tuning, \texttt{Stage III}: Mask-Text Instruction Fine-tuning.
}
\label{tab:details}
\vspace{-10pt}
\begin{center}
\resizebox{\linewidth}{!}{
\begin{tabular}{lccc}
 & \texttt{Stage I} & \texttt{Stage II} & \texttt{Stage III}\\
 \toprule[0.95pt]
 Phase & Image-Text Alignment & Image-Text SFT & Mask-Text SFT\\
 \midrule[0.6pt]
 \multicolumn{4}{c}{\textit{Training Hyper-Parameters}} \\
 \midrule[0.6pt]
 Input image resolution & 256$\times$256 & 448$\times$448 & 448$\times$448 \\
 Output image resolution & 512$\times$512 & 512$\times$512 & 512$\times$512 \\
 VFM & CLIP-ConvNext-B & CLIP-ConvNext-B & CLIP-ConvNext-B\\
 LLM & Vicuna-7B v1.5 & Vicuna-7B v1.5 & Vicuna-7B v1.5\\
 DM & Stable Diffusion v2.1 & Stable Diffusion v2.1 & Stable Diffusion v2.1\\
 $\lambda$ & 5 & 5 & 5\\
 \multirow{2}{*}[-0.0ex]{Learning Rate} & 2e-5 (image encoder\&decoder) & 1e-6 (language model) & 1e-6 (language model)\\
 & 1e-4 (others) & 1e-5 (others) & 1e-5 (others)\\
Optimizer & AdamW & AdamW & AdamW\\
 Optimizer hyper-parameters & $\beta_1,\beta_2,\epsilon=0.9,0.995,$1e-6 & $\beta_1,\beta_2,\epsilon=0.9,0.999,$1e-8 & $\beta_1,\beta_2,\epsilon=0.9,0.999,$1e-8 \\
 Weight Decay & 0.05 & 0.05 & 0.05\\
 Training iterations & 10k & 10k & 10k\\
 Warmup steps & 1k & 500 & 500 \\
 Learning Rate Scheduler & Cosine & Cosine & Cosine\\
 Batch Size Per GPU & 4 & 16 & 2\\
 Maximum Token Length & 2048 & 2048 & 2048\\
 Augmentation & CenterCrop & - & -\\
 Unfreeze LLM & \xmark & \cmark & \cmark \\
  Unfreeze DM & \xmark & \xmark & \xmark \\
  Unfreeze VFM & \cmark & \xmark & \xmark \\
 \midrule[0.6pt]
 \multicolumn{4}{c}{\textit{Training Data}} \\
 \midrule[0.6pt]
 \multirow{4}{*}[-0.0ex]{Dataset} 
 & \ding{192} MMC4  & \ding{192} LLaVA-Mix-665K & \ding{192} COCO/ReferCOCO/ReferCOCO+\\
 & \ding{193} BLIP-LCS  & \ding{193} VQA-Mixture & \ding{193} Pascal-Part/Part-ImageNet\\
 &   & \ding{194} COCO Caption & \ding{195} VG/VRC\\
 &  &  & \\
 \midrule[0.3pt]
 Data Size & $\sim$26M & $\sim$800K & $\sim$200K \\
 Data Type & Interleave/Pair & Instruction & Instruction\\
  \midrule[0.6pt]
 \multicolumn{4}{c}{\textit{Training Cost}} \\
 \midrule[0.6pt]
 GPU Device & 32$\times$NVIDIA A100 & 32$\times$NVIDIA A100 & 32$\times$NVIDIA A100\\ 
 Training Time & $\sim$30h & $\sim$6h & $\sim$3h\\
\bottomrule[0.95pt]
\end{tabular}
}
\end{center}
\end{table*}

\begin{table}[t]
\centering
\caption{\textbf{Overall descriptions of the evaluation benchmarks} for evaluating capabilities, including image-level captioning, image-level visual question answering, text-to-image generation, region-level image captioning, visual robustness, comprehension, perception and hallucination.}\label{tab:eval_benchmarks_summary}
\small
\resizebox{\textwidth}{!}{%
\begin{tabular}{@{}lllll@{}}
\toprule[0.95pt]
& Dataset & Task description & Eval Split & Metric \\
\cmidrule(l){1-5} 
\multirow{2}{*}{\rotatebox[origin=c]{90}{CAP.}} 
& COCO~\citep{coco}  & Scene description   & \texttt{test} &  CIDEr($\uparrow$)~\citep{cider} \\
& Image2Paragraph~\citep{img2paragraph}    & Scene description   & \texttt{test} &  CIDEr($\uparrow$)~\citep{cider} \\
\midrule[0.6pt]
 \multirow{5}{*}{\rotatebox[origin=c]{90}{VQA.}} & VQAv2~\citep{vqav2}        & Scene understanding QA & \texttt{test-dev}  &  VQA Acc($\uparrow$)~\citep{vqa-acc}     \\
& OKVQA~\citep{ok-vqa}         & External knowledge QA &  \texttt{val}  &  VQA Acc($\uparrow$)~\citep{vqa-acc}     \\
& GQA~\citep{gqa}         & Scene understanding QA &  \texttt{test-dev}  &  VQA Acc($\uparrow$)~\citep{vqa-acc}     \\
& VizWiz~\citep{vizwiz}      & Scene understanding QA &  \texttt{test-dev}   &  VQA Acc($\uparrow$)~\citep{vqa-acc}     \\
& VisDial~\citep{visdial}      &  Image dialogue  &  val  & NDCG($\uparrow$)   \\
\midrule[0.6pt]
 \multirow{2}{*}{\rotatebox[origin=c]{90}{\small{SYN.}}} & MS-COCO ~\citep{mscoco} & Text-Conditional Image Synthesis  &  \texttt{val}-30K   & FID($\downarrow$)~\citep{fid}\\
 & LN-COCO~\citep{lncoco} & Text-Conditional Image Synthesis  & \texttt{val} & FID($\downarrow$)~\citep{fid}\\
\midrule[0.6pt]
\multirow{3}{*}{\rotatebox[origin=c]{90}{REF.}}& RefCOCO~\citep{refercoco} & Region-level scene description & \texttt{val} & CIDEr($\uparrow$)~\citep{cider}\\
& RefCOCO+~\citep{refercoco+g} &  Region-level scene description &  \texttt{val}   & CIDEr($\uparrow$)~\citep{cider}\\
& RefCOCOg~\citep{refercoco+g} & Region-level scene description & \texttt{val} & CIDEr($\uparrow$)~\citep{cider}\\
\midrule[0.6pt]
\multirow{3}{*}{\rotatebox[origin=c]{90}{OOD.}}

& RobustVQA-V & Out-of-Distribution Robustness & \texttt{val}& Acc($\uparrow$)\\
& RobustVQA-R & Out-of-Distribution Robustness & \texttt{val} & Acc($\uparrow$)\\
& RobustVQA-A & Out-of-Distribution Robustness & \texttt{val} & Acc($\uparrow$)\\

\midrule[0.6pt]
\multirow{3}{*}{\rotatebox[origin=c]{90}{Hall.}}
& POPE-R~\citep{pope} & Visual Hallucination & \texttt{val}& Acc($\uparrow$)\\
& POPE-P~\citep{pope} & Visual Hallucination & \texttt{val} & Acc($\uparrow$)\\
& POPE-A~\citep{pope} & Visual Hallucination & \texttt{val} & Acc($\uparrow$)\\
\midrule[0.6pt]
\multirow{2}{*}{\rotatebox[origin=c]{90}{CPH.}}
& MMBench~\citep{mmvet} & Visual Comprehension & \texttt{val}& Acc($\uparrow$)\\
& MMVet~\citep{mmvet} & Visual Comprehension & \texttt{val} & Acc($\uparrow$)\\
\midrule[0.6pt]
\rotatebox[origin=c]{90}{PCP.}
& MMVP~\citep{mmvp} & Visual Perception & \texttt{val}& Acc($\uparrow$)\\
\bottomrule[0.95pt]
\end{tabular}
}
\end{table}

\section{Additional Visualization Examples}\label{app:add_vis_example}
\subsection{Semantic Image Synthesis}

\begin{figure*}[h]
    \centering
    \includegraphics[width=0.85\linewidth]{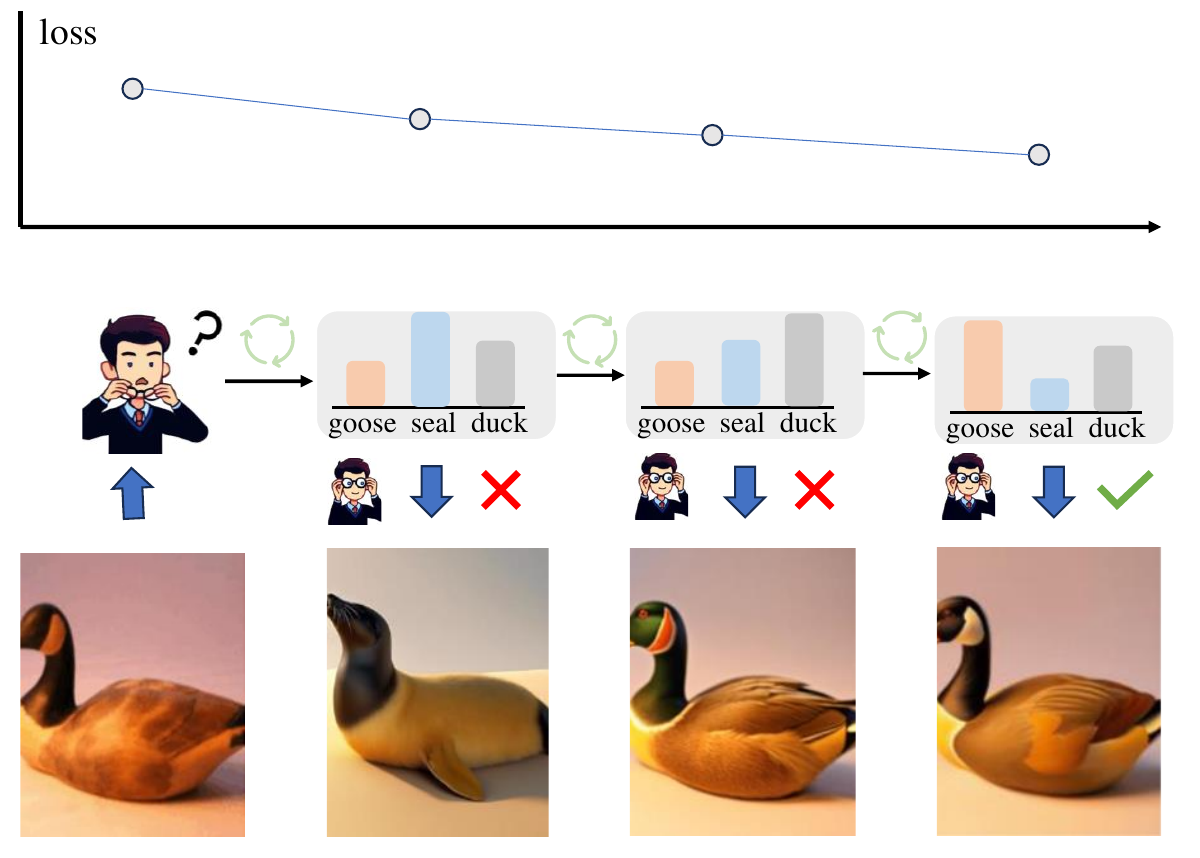}
    \caption{\textbf{Dynamic semantic bias elimination process} through three iterations on the same sample, diffusion process is conducted by adding 65\% noise to the original image as the initial condition.}
\label{fig:dynamic_semantic_bias}
\end{figure*}

\noindent \textbf{Dynamic Semantic Bias Erasure.
We demonstrate the dynamic semantic bias elimination process through three iterations on the same sample, providing an illustration of the original image alongside its version reconstructed in real-time according to semantic conditions, as shown in \cref{fig:dynamic_semantic_bias}. Our method, \name, gradually mitigates potential erroneous semantics within the visual encoder through multiple iterations, ultimately enhancing the perceptual capabilities of MLLMs.}

\begin{figure*}[h]
    \centering
    \includegraphics[width=\linewidth]{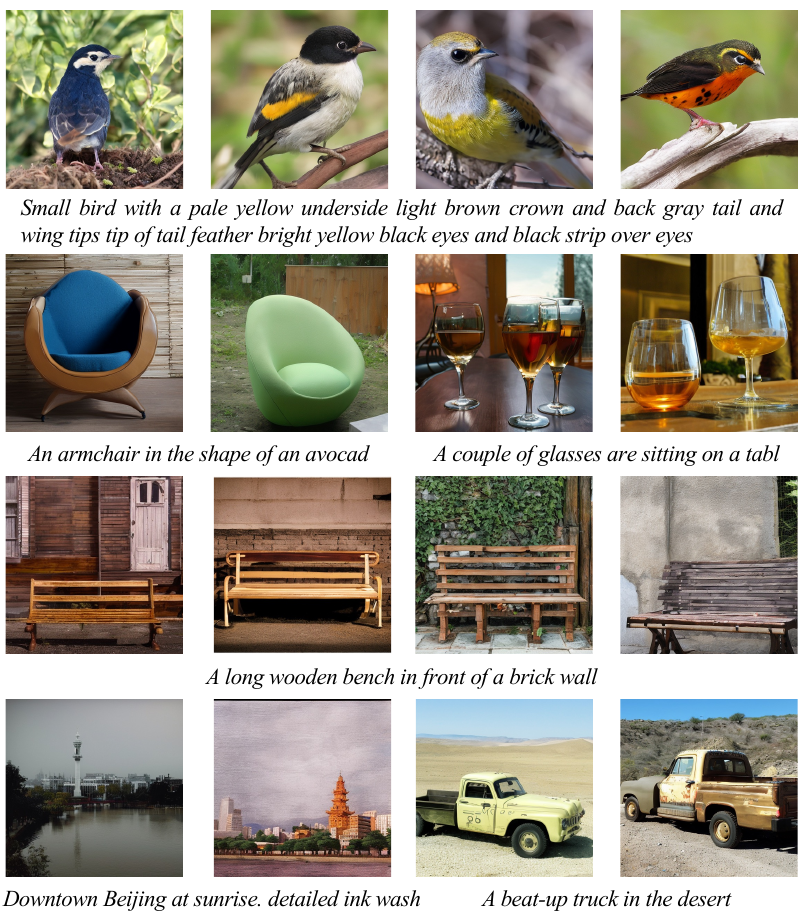}
    \caption{\textbf{Text-to-image generation examples with prompts.} \name can generate vivid images based on input text conditions.}
\label{fig:image_to_image_case}
\end{figure*}

\noindent \textbf{Consistency Semantic Image Synthesis}
We visualize some consistency semantic image synthesis and display both the original images and their reconstructed versions in \cref{fig:image_to_image_case}. \name accurately recovers the features of the original images without causing distortion.
\begin{figure*}[h]
    \centering
    \includegraphics[width=\linewidth]{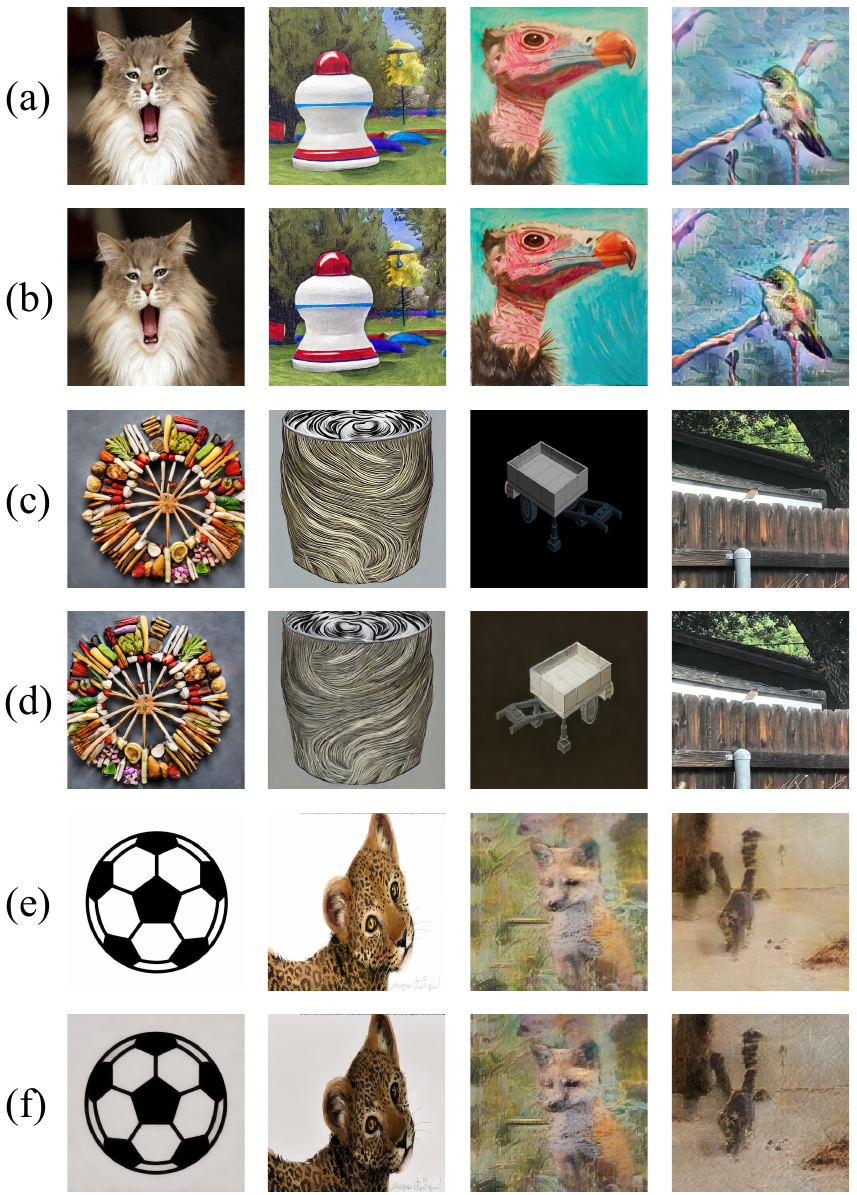}
    \caption{\textbf{image-to-image generation examples with the outputs of image encoder.} (a,c,e) are original images and (b,d,f) are synthesis images based on the image embeddings of original images.}
\label{fig:text_to_image_case}
\end{figure*}
\subsection{Text Condition Image Synthesis}
In \cref{fig:text_to_image_case}, we present some text-to-image synthesis examples from \name, demonstrating its capability to generate corresponding images based on given prompts. 

\subsection{Robustness Comparison}
In \cref{fig:case_comparison}, we present a comparative analysis of visual robustness results between our model, \name, and other state-of-the-art models: LLaVA~\citep{llava}, NeXt-Chat~\citep{nextchat}, and MM-Interleaved~\citep{mm-interleaved}. When encountering natural adversarial samples or out-of-distribution samples, the image encoder in their models will output incorrect semantic information, leading to incorrect category answers. In contrast, our method uses a diffusion model as the eyes of the large language model to inspect and correct the output features of the image encoder. This process eliminates incorrect semantic outputs from the image encoder, ultimately allowing the large language model to produce the correct category answer. This simple yet effective approach significantly enhances the model's robustness and generalization capabilities.

\subsection{Image-Text Multimodal Dialogue}
\begin{figure*}[h]
    \centering
    \includegraphics[width=0.86\linewidth]{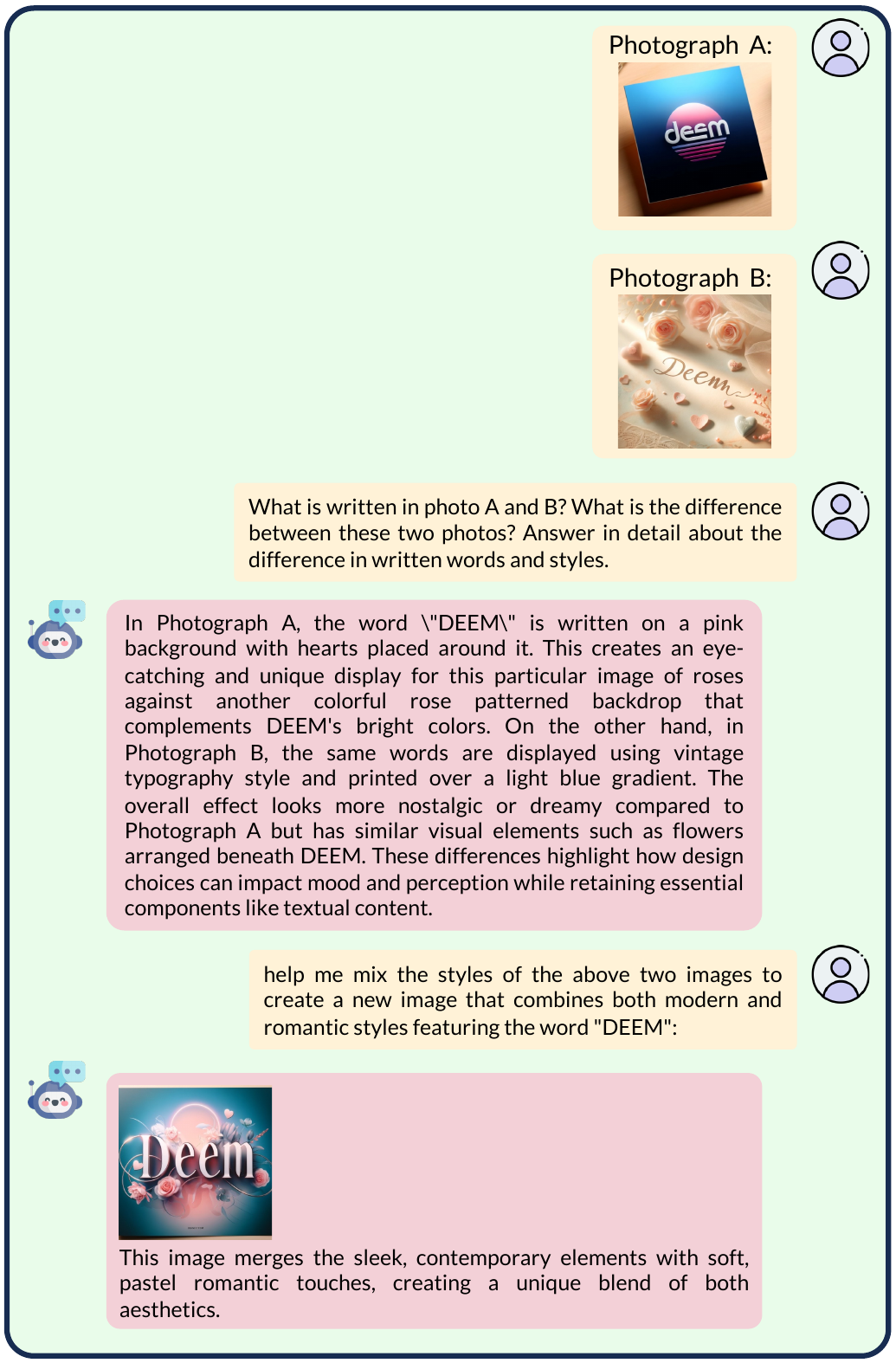}
    \caption{\textbf{Examples of image-text multimodal dialogue} between human and \name. Text and image can be used as inputs or outputs, and multi-round dialogue is shown.}
\label{fig:img_txt_case}
\end{figure*}

In \cref{fig:img_txt_case}, we show the image-text dialogue case examples of \name. Our model can input any interleaved layout of text-image data and simultaneously understand and generate text-image outputs in any interleaved layout, representing the future of next-generation multimodal dialogue.
\subsection{Mask-Text Multimodal Dialogue}
\begin{figure*}[h]
    \centering
    \includegraphics[width=\linewidth]{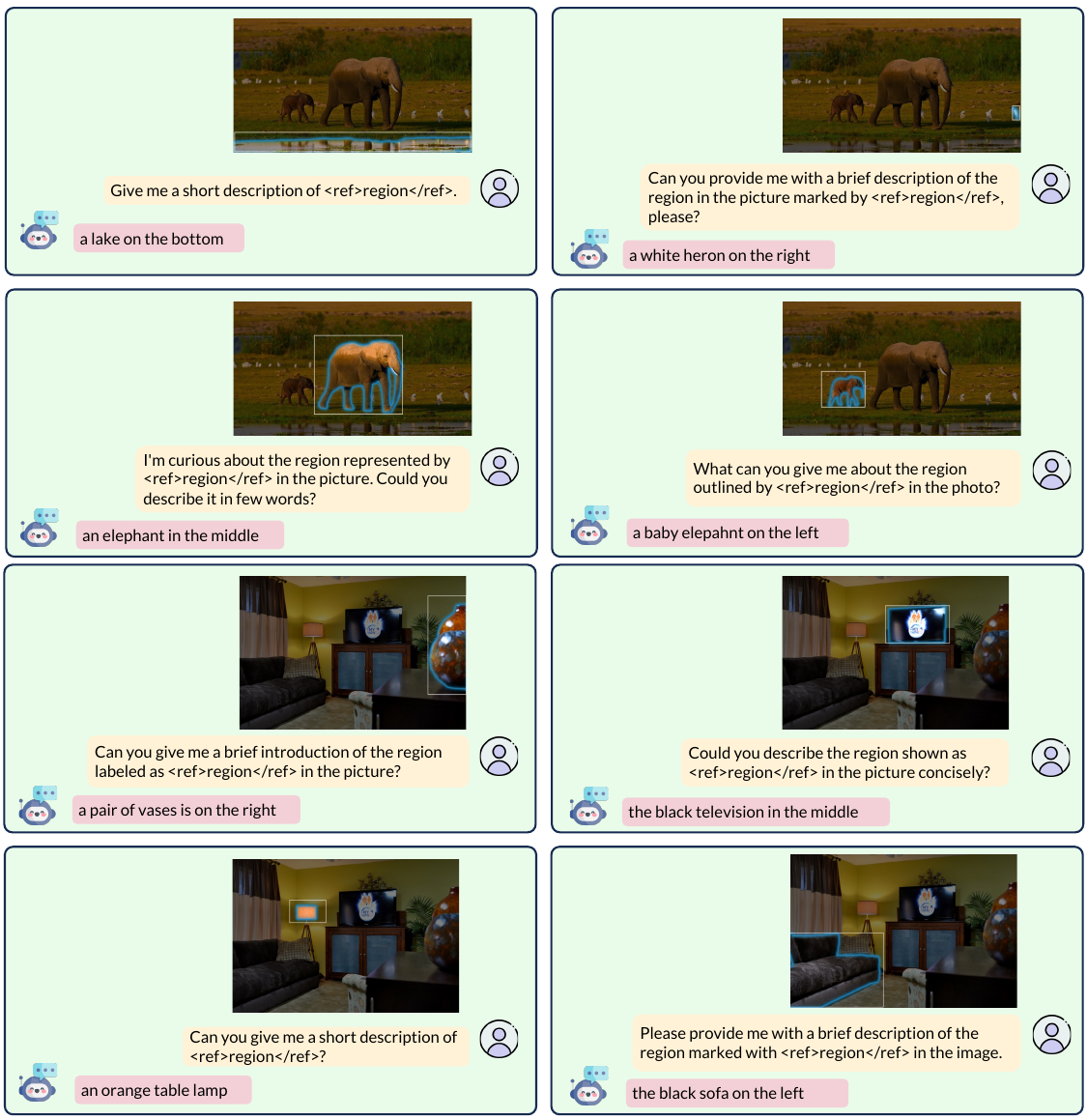}
    \caption{\textbf{Examples of mask-text multimodal dialogue} between human and \name. Text and mask can be used as inputs and \name outputs the corresponding answer, and multi-round dialogue is shown.}
\label{fig:mask_txt_case}
\end{figure*}

In addition to image-level input, \name also supports mask-text input to perform fine-grained region-level reasoning tasks. As shown in the \cref{fig:mask_txt_case}, \name can accurately extract region semantics of the image based on the input mask and complete the corresponding instruction tasks.

\end{document}